\documentclass[runningheads]{llncs}

\usepackage{eccv}

\usepackage{eccvabbrv}

\usepackage{graphicx}
\usepackage{booktabs}
\usepackage{multicol}
\usepackage{multirow}
\usepackage{arydshln}
\usepackage{xurl}

\usepackage[accsupp]{axessibility}  

\usepackage[pagebackref,breaklinks,colorlinks=true,citecolor=MidnightBlue,linkcolor=MidnightBlue]{hyperref}

\usepackage{orcidlink}

\begin{document}

\title{Zero-Shot Multi-Object Scene Completion}


\author{
Shun Iwase\inst{1,2} \and
Katherine Liu\inst{2} \and
Vitor Guizilini\inst{2} \and
Adrien Gaidon\inst{2} \and \\
Kris Kitani\inst{1,\star} \and
Rareș Ambruș\inst{2,\star} \and
Sergey Zakharov\inst{2,}\thanks{Equal advising.}
}

\authorrunning{S.~Iwase et al.}

\institute{Carnegie Mellon University \and Toyota Research Institute}

\maketitle

\begin{center}
    \centering
    \includegraphics[width=\textwidth]{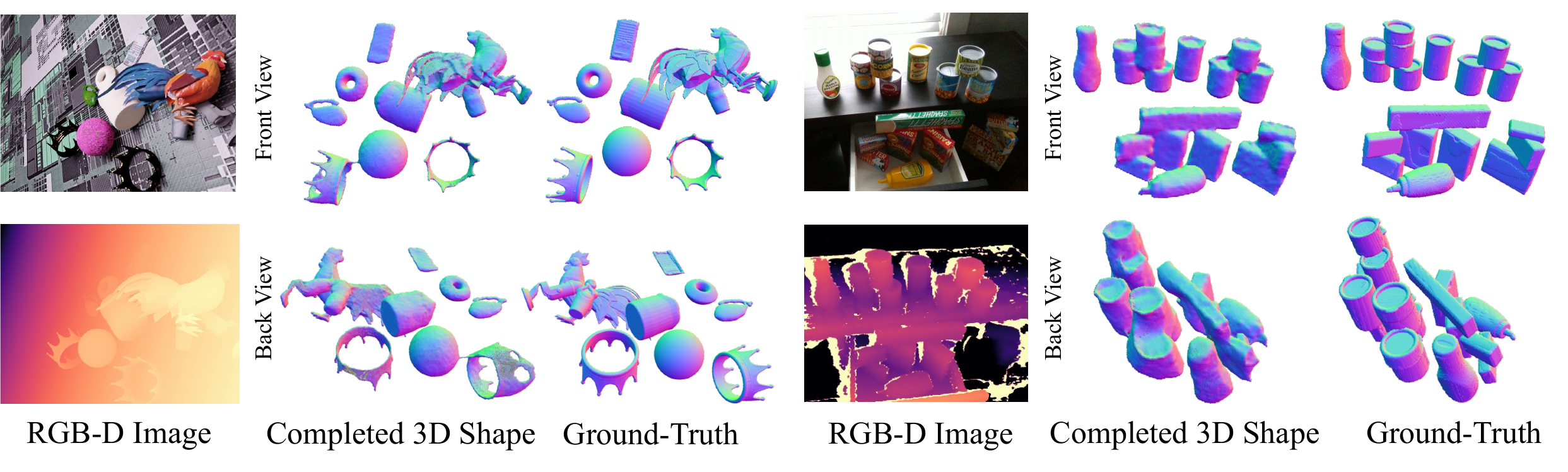}
    \captionof{figure}{Given an RGB-D image and the foreground mask of multiple objects not seen during training, our method predicts their complete 4D shapes quickly and accurately, including occluded areas. (\textbf{Left}) Synthetic image results. (\textbf{Right}) Zero-shot generalization to a real-world image of household objects with noisy depth data.
    Our 3D results are rotated with respect to the input to highlight completions in occluded regions.
}
    \label{fig:teaser}
\end{center}%

\begin{abstract}
We present a 3D scene completion method that recovers the complete geometry of multiple unseen objects in complex scenes from a single RGB-D image. Despite notable advancements in single-object 3D shape completion, high-quality reconstructions in highly cluttered real-world multi-object scenes remains a challenge.
To address this issue, we propose OctMAE, an architecture that leverages an Octree U-Net and a latent 3D MAE to achieve high-quality and near real-time multi-object scene completion through both local and global geometric reasoning.
Because a naive 3D MAE can be computationally intractable and memory intensive even in the latent space, we introduce a novel occlusion masking strategy and adopt 3D rotary embeddings, which significantly improve the runtime and scene completion quality.
To generalize to a wide range of objects in diverse scenes, we create a large-scale photorealistic dataset, featuring a diverse set of $12$K 3D object models from the Objaverse dataset that are rendered in multi-object scenes with physics-based positioning. Our method outperforms the current state-of-the-art on both synthetic and real-world datasets and demonstrates a strong zero-shot capability. \url{https://sh8.io/#/oct_mae}

\end{abstract}

\begin{figure*}[t!]
    \centering
    \includegraphics[width=\textwidth]{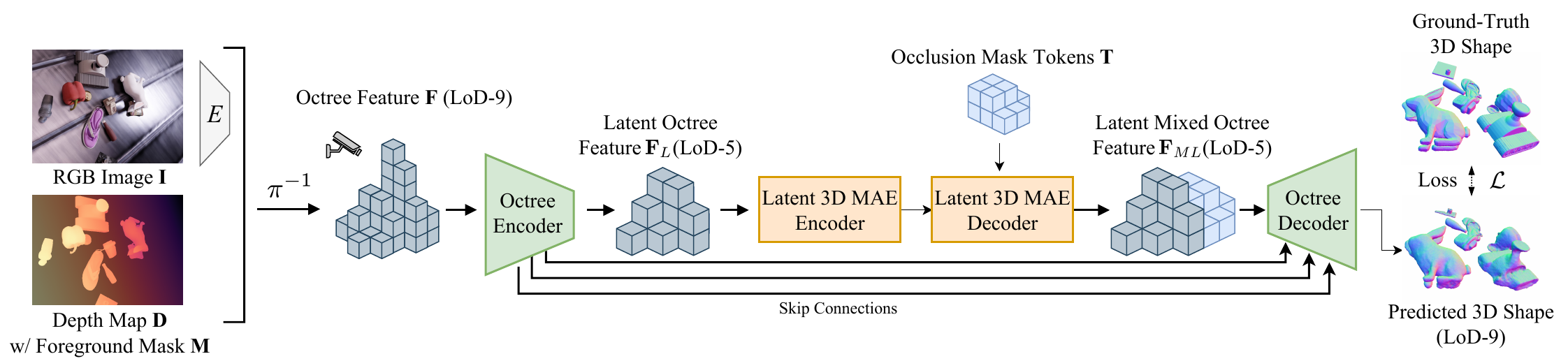}
    \caption{Overview of our proposed method (OctMAE). Given an input RGB Image $\mathbf{I}$, depth map $\mathbf{D}$, and a foreground mask $\mathbf{M}$, the octree feature $\mathbf{F}$ is obtained by unprojecting an image feature encoded by a pre-trained image encoder $\mathbf{E}$. The octree feature is then encoded by the Octree encoder and downsampled to the Level of Detail (LoD) of $5$. The notation LoD-$h$ indicates that each axis of the voxel grid has resolution of $2^h$. The latent 3D MAE takes the encoded Octree feature $\mathbf{F}$ as input and its output feature is concatenated with the occlusion mask tokens $\mathbf{T}$. Next, the masked decoded feature $\mathbf{F}_{ML}$ is computed by sparse 3D MAE decoder. Finally, the Octree decoder predicts a completed surface at LoD-$9$.}
    \label{fig:overview}
\end{figure*}

\section{Introduction}
\label{sec:intro}

Humans can instantly imagine complete shapes of multiple novel objects in a cluttered scene via advanced geometric and semantic reasoning. This ability is also essential for robots if they are to effectively perform useful tasks in the real world~\cite{2017_rss_system,varley2017shape,10.1177/0278364911406761,okumura2023quick}. In this work, we propose a method that can quickly and accurately complete a wide number of objects in diverse real-world scenes. 

Prior works~\cite{yan2022shapeformer,autosdf2022,Park_2019_CVPR,li2023voxformer,Li2020aicnet,lin2021fusion} have achieved phenomenal progress in scene and object shape completion from a single RGB-D image. Object-centric methods~\cite{duan2020curriculum,irshad2022shapo} in particular can achieve very high reconstruction accuracy by relying on category-specific shape priors. However, when deployed on entire scenes such methods require bespoke instance detection/segmentation models, and often perform test-time optimization which is time consuming and would hinder real-time deployment on a robot. Moreover, existing methods are typically limited to a small set of categories. Thus, zero-shot multi-object scene completion remains a challenging and open problem that has seen little success to date.
This is in stark contrast to the sudden increase in powerful algorithms for 2D computer vision tasks such as object detection~\cite{zhang2022glipv2,li2021grounded} and image segmentation~\cite{xu2022odise,liang2023open}. We attribute this progress to a great extent to the availability of large-scale datasets~\cite{schuhmann2022laion,together2023redpajama} coupled with neural architectures and learning objectives~\cite{radford2021learning,rombach2021highresolution,he2022masked,shi2020improving} that can effectively exploit the highly structured data occurring in the natural world~\cite{goldblum2023no}. 

Taking inspiration from the latest developments in the 2D domain, we propose a scene completion algorithm at the scene level that generalizes across a large number of shapes and that only supposes an RGB-D image and foreground mask as input. Our method consists of Octree masked autoencoders (OctMAE) --- a hybrid architecture of Octree U-Net and a latent 3D MAE (\Cref{fig:overview}). Although a recent work, VoxFormer~\cite{li2023voxformer}, also extends MAE architecture to 3D using deformable 3D attention and shows great improvement in semantic scene completion tasks, its memory utilization is still prohibitive to handle a higher resolution voxel grid. We address this issue by integrating 3D MAE into the latent space of Octree U-Net. Our experiments show that the latent 3D MAE is the key to global structure understanding and leads to strong performance and generalization across all datasets. Moreover, we find that the choice of a masking strategy and 3D positional embeddings is crucial to achieve better performance. We provide extensive ablations to verify that our 3D latent MAE design is effective.

Our second contribution consists of the creation of a novel synthetic dataset to counteract the lack of large-scale and diverse 3D datasets. The dataset contains $12$K 3D models of hand-held objects from Objaverse~\cite{objaverse} and GSO~\cite{10.1109/ICRA46639.2022.9811809} datasets (\Cref{fig:dataset_example}). We utilize the dataset to conduct a comprehensive evaluation of our method as well as other baselines and show that our method scales and achieves better results. Finally, we perform zero-shot evaluations on synthetic as well as real datasets and show that a combination of 3D diversity coupled with an appropriate architecture is key to generalizable scene completion in the wild. 

Our contributions can be summarized as follows:

\begin{itemize}
    \item We present a novel network architecture, Octree Masked Autoencoders (OctMAE), a hybrid architecture of Octree U-Net and latent 3D MAE, which achieves state-of-the-art results on all the benchmarks. Further, we introduce a simple occlusion masking strategy with full attention, which boosts the performance of a latent 3D MAE.
    \item We create the first large-scale and diverse synthetic dataset using Objaverse~\cite{objaverse} dataset for zero-shot multi-object scene completion, and provide a wide range of benchmark and analysis.
\end{itemize}

\section{Related Work}
\label{sec:related_work}

\paragraph{\textbf{3D reconstruction and completion.}}
Reconstructing indoor scenes and objects from a noisy point cloud has been widely explored \cite{OccupancyNetworks,Peng2020ECCV,hou2020revealnet,dai2020sgnn,bozic2021transformerfusion,dai2018scancomplete,10.1007/978-3-031-19824-3_30,choy20194d,williams2021nkf,huang2023nksr,li2023voxformer,wu2023multiview,Boulch_2022_CVPR,shen2021dmtet,Liu2023MeshDiffusion,Park_2019_CVPR}. 
Several works \cite{yu2021pointr,yan2022shapeformer,autosdf2022,cheng2023sdfusion,varley2017shape,PSWang2020,song2016ssc,zhang2022cgca,10160350,10.1007/978-3-031-19824-3_30,Zhang_2019_ICCV,Park_2019_CVPR} tackle more challenging shape completion tasks where large parts of a target is missing. While these methods achieve impressive results, they do not explicitly consider semantic information, which may limit their capability for accurate shape completion. Recent methods~\cite{Li2020aicnet,li2023voxformer,Zhang_2019_ICCV,Li2019ddr} in Semantic Scene Completion (SSC) leverage semantic information via an RGB image. Nevertheless, the number of target categories is quite limited, restricting its utility for a broad range of applications in the real world. In addition, many methods adopt occupancy or SDF as an output representation, which necessitates post-processing such as the marching cubes~\cite{10.1145/37402.37422} and sphere tracing to extract an explicit surface. As another direction, GeNVS\cite{chan2023genvs}, Zero-1-to-3~\cite{liu2023zero1to3}, and 3DiM~\cite{Watson2022NovelVS} explore single-view 3D reconstruction via novel view synthesis. However, expensive test-time optimization is required. Recently, One-2-3-45~\cite{liu2023one} and MCC~\cite{wu2023multiview} attempt to improve the generation speed, however, their runtime for multi-object scenes is still far from near real-time. Further, since these methods are object-centric, multiple objects in a single scene are not handled well due to the complicated geometric reasoning especially caused by occlusions by other objects. In this paper, we propose a general and near real-time framework for multi-object 3D scene completion in the wild using only an RGB-D image and foreground mask without expensive test-time optimization.

\paragraph{\textbf{Implicit 3D representations.}}
Recently, various types of implicit 3D representation have become popular in 3D reconstruction and completion tasks. Early works~\cite{Park_2019_CVPR,OccupancyNetworks,functa22} use a one-dimensional latent feature to represent a 3D shape as occupancy and SDF fields. Several works~\cite{Peng2020ECCV,Li2020aicnet,song2016ssc} employ voxels, ground-planes, and triplanes, demonstrating that the retention of geometric information using 3D CNNs enhances performance. Although the voxel representation typically performs well among these three, its cubic memory and computational costs make increasing resolution challenging. To mitigate this issue, sparse voxels~\cite{10.1145/3072959.3073608,choy20194d,8580422,3DSemanticSegmentationWithSubmanifoldSparseConvNet,liu2020neural} treat a 3D representation as a sparse set of structured points using the octree and hash table and perform convolutions only on non-empty voxels and its neighbors. Further, the high-resolution sparse voxel enables a direct prediction of a target surface. As another direction, \cite{Boulch_2022_CVPR,zhao2021point,wu2022point} leverage point cloud. Nonetheless, an unstructured set of points can be non-uniformly distributed in the 3D space and requires running the k-NN algorithm at every operation. This aspect often renders point-based methods less appealing compared to the sparse voxel representation. Therefore, our method adopts an octree-based representation used in \cite{10.1145/3072959.3073608} for efficient training and direct surface prediction.

\paragraph{\textbf{Masked Autoencoders (MAE).}}
Inspired by the success of ViTs~\cite{dosovitskiy2020vit,zhai2021scaling} and masked language modeling~\cite{Devlin2019BERTPO,Radford2018ImprovingLU}, \cite{he2022masked} demonstrates that masked autoencoders (MAE) with ViTs can learn powerful image representation by reconstructing masked images. To improve the efficiency and performance of MAE, ConvMAE~\cite{gao2022convmae} proposes a hybrid approach that performs masked autoencoding at the latent space of 2D CNN-based autoencoder network. Recently, VoxFormer~\cite{li2023voxformer} extends the MAE design to 3D for semantic scene completion using 3D deformable attention, and shows great improvement over previous works. However, it is not trivial to scale up the MAE architecture to a higher resolution voxel due to memory constraints. Motivated by ConvMAE~\cite{gao2022convmae} and OCNN~\cite{10.1145/3072959.3073608}, we propose an efficient OctMAE architecture using sparse 3D operations.

\section{Proposed Method}
\label{sec:method}

Given an RGB image $\mathbf{I} \in \mathbb{R}^{H \times W \times 3}$, depth map $\mathbf{D} \in \mathbb{R}^{H \times W}$, and foreground mask $\mathbf{M} \in \mathbb{R}^{H \times W}$ containing all objects of interest, we aim to predict their complete 3D shapes quickly and accurately.
Our framework first encodes an RGB image $\mathbf{I}$ with a pre-trained image encoder $E$ such as ResNeXt~\cite{Xie2016} and then lifts the resulting features up to 3D space using a depth map $\mathbf{D}$ and foreground mask $\mathbf{M}$ to acquire 3D point cloud features $\mathbf{F} \in \mathbb{R}^{N \times D}$ and its locations $\mathbf{P} \in \mathbb{R}^{N \times 3}$ (\Cref{subsec:point_cloud_feature_extraction}). Second, we convert the 3D features into an octree using the same algorithm used in \cite{PSWang2020} and pass it to OctMAE to predict a surface at each LoD (\Cref{subsec:3d_sparse_mae}). The diagram of our method is visualized in \Cref{fig:overview}.

\subsection{Octree Feature Aggregation}
\label{subsec:point_cloud_feature_extraction}
We adopt ResNeXt-50~\cite{Xie2016} as an image encoder to obtain dense and robust image features $\mathbf{W} = E\left(\mathbf{I}\right) \in \mathbb{R}^{H \times W \times D}$ from an RGB image. The image features are unprojected into the 3D space using a depth image with $\left(\mathbf{F}, \mathbf{P}\right) = \pi^{-1}\left(\mathbf{W}, \mathbf{D}, \mathbf{M}, \mathbf{K} \right)$ where a point cloud feature and its corresponding coordinates are represented as $\mathbf{F}$ and $\mathbf{P}$. $\pi^{-1}$ unprojects the image features $\mathbf{W}$ to the camera coordinate system using a depth map $\mathbf{D}$, foreground mask $\mathbf{M}$, and an intrinsic matrix $\mathbf{K}$. Next, we define an octree at the level of detail (LoD) of $9$ ($512^3$) with the grid and cell size being $1.28$ m and $2.5$ mm respectively, and use the point features to populate the voxel grid, averaging features when multiple points fall into the same voxel. Here, LoD-$h$ simply represents resolution of an octree. For instance, the voxel grid of LoD-$9$ has the maximum dimension of $2^9 = 512$ for each axis. An octree is represented as a set of $8$ octants with features at non-empty regions; therefore, it is more memory-efficient than a dense voxel grid. The octree is centered around the z-axis in the camera coordinate system, and its front plane is aligned with the nearest point to the camera along with the z-axis.

\subsection{OctMAE: Octree Masked Autoencoders}
\label{subsec:3d_sparse_mae}
We design OctMAE which leverages Octree U-Net~\cite{10.1145/3072959.3073608} and latent 3D MAE to achieve accurate and efficient zero-shot multi-object scene completion.
Octree U-Net consists of multiple sparse 3D convolutional layers. While the Octree U-Net architecture can efficiently encode octree features to low resolution, only local regions are considered at each operation. On the contrary, 3D MAE can capture global object information which helps predict globally consistent 3D shapes. However, unlike an image, a dense voxel grid contains a prohibitive number of tokens even in the latent space, which makes it challenging to adopt an MAE architecture directly for 3D tasks. Recently, ConvMAE~\cite{gao2022convmae} proposed to leverage the advantages of both CNNs and MAE in 2D for efficient training. Nevertheless, a na\"ive extension of ConvMAE~\cite{gao2022convmae} to 3D also leads to prohibitive computational and memory costs. To address this issue, we propose a novel occlusion masking strategy and adopt 3D rotary embeddings, enabling efficient masked autoencoding in the latent space.

\paragraph{\textbf{Encoder architecture}.}
The encoder of Octree U-Net~\cite{PSWang2020} takes the octree feature at LoD-$9$ and computes a latent octree feature $\mathbf{F}_{L} \in \mathbb{R}^{N^\prime \times D^\prime}$ at LoD-$5$ where $N^\prime$ is the number of non-empty voxels and $D^\prime$ is the latent feature dimension.
To incorporate global symmetric and object scale information which gives more cues about completed shapes, we use $S$ layers of the full self-attention Transformer blocks in the latent 3D MAE encoder. Since $N^\prime$ is typically the order of the hundreds to thousands, we resort to memory-efficient attention algorithms~\cite{rabe2021selfattention,dao2023flashattention2}. Ideally, learnable relative positional encodings~\cite{zhao2021point} are used to deal with the different alignments of point cloud features inside an octree. However, it requires computing the one-to-one relative positional encoding $N^\prime \times N^\prime$ times, which largely slows down the training and makes it computationally impractical. Therefore, we use RoPE~\cite{su2021roformer} to encode 3D axial information between voxels. Concretely, we embed position information with RoPE at every multi-head attention layer as

\begin{equation}
    \mathbf{R}_i = \mathop{\mathrm{diag}}\left(R (p^x_i), R (p^y_i), R (p^z_i), \mathbf{I} \right) \in \mathbb{R}^{D^\prime \times D^\prime}, \quad
    \mathbf{f}^{\prime}_i = \mathbf{R}_i \mathbf{f}_i,
\end{equation}
where $\mathbf{f}_i \in \mathbb{R}^{D^\prime}$, and $\mathbf{p}_i \in \mathbb{R}^{3}$ is $i$-th octree feature and coordinates. $R: \mathbb{R} \rightarrow \mathbb{R}^{\left\lfloor{D^{\prime} / 3}\right\rfloor \times \left\lfloor{D^{\prime} / 3}\right\rfloor}$ is a function to generate a rotation matrix given normalized 1D axial coordinate. The detailed derivation of $\mathbf{R}$ can be found in the supplemental.

\begin{figure}[t!]
\begin{minipage}{.6\textwidth}
    \centering
    \begin{subfigure}[t]{0.32\textwidth}
    \includegraphics[width=\textwidth]{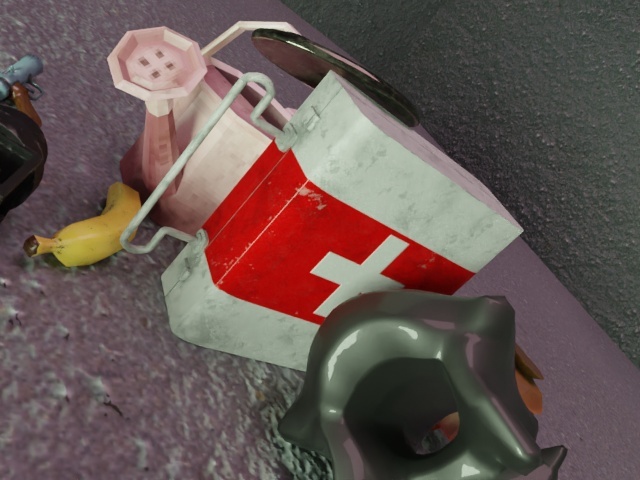}%
    \end{subfigure}
    \begin{subfigure}[t]{0.32\textwidth}
    \includegraphics[width=\textwidth]{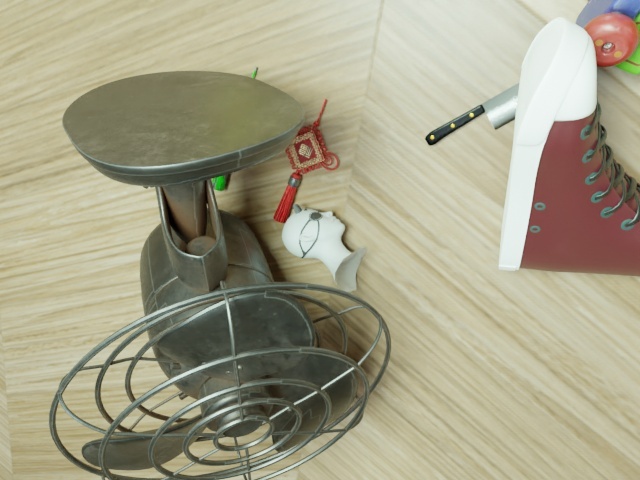}
    \end{subfigure}
    \begin{subfigure}[t]{0.32\textwidth}
    \includegraphics[width=\textwidth]{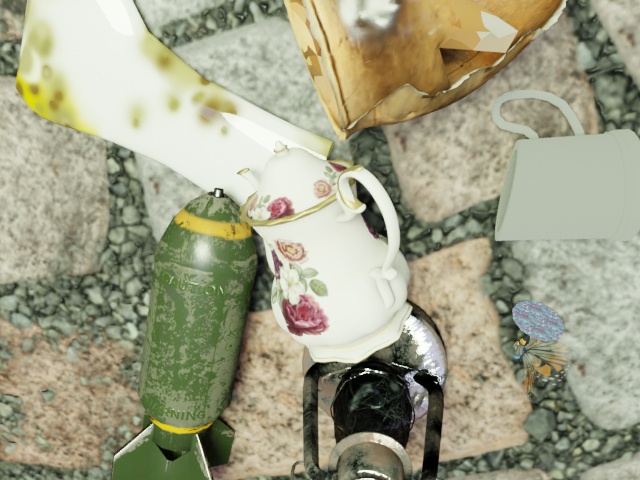}
    \end{subfigure} \\
    \begin{subfigure}[t]{0.32\textwidth}
    \includegraphics[width=\textwidth]{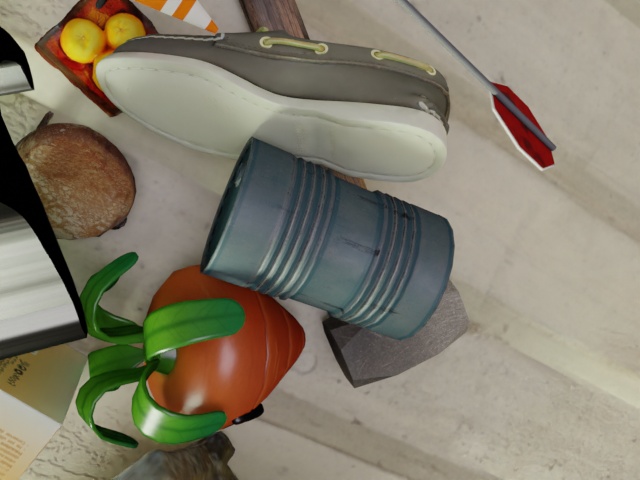}
    \end{subfigure}
    \begin{subfigure}[t]{0.32\textwidth}
    \includegraphics[width=\textwidth]{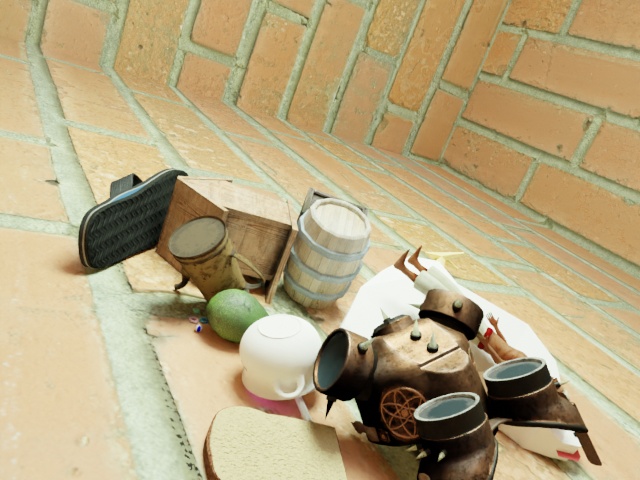}%
    \end{subfigure}
    \begin{subfigure}[t]{0.32\textwidth}
    \includegraphics[width=\textwidth]{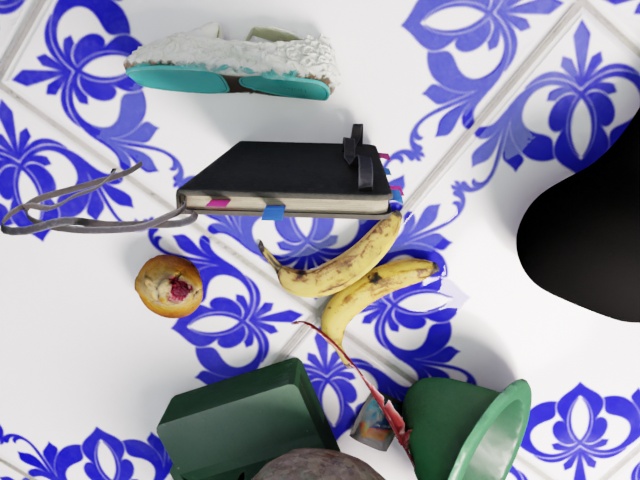}%
    \end{subfigure}
    \caption{Example images of our synthetic dataset. We use BlenderProc~\cite{Denninger2023} to acquire high-quality images under various and realistic illumination conditions.}
    \label{fig:dataset_example}
    \end{minipage} \quad
    \begin{minipage}{0.35\textwidth}
        \centering
        \includegraphics[width=\linewidth]{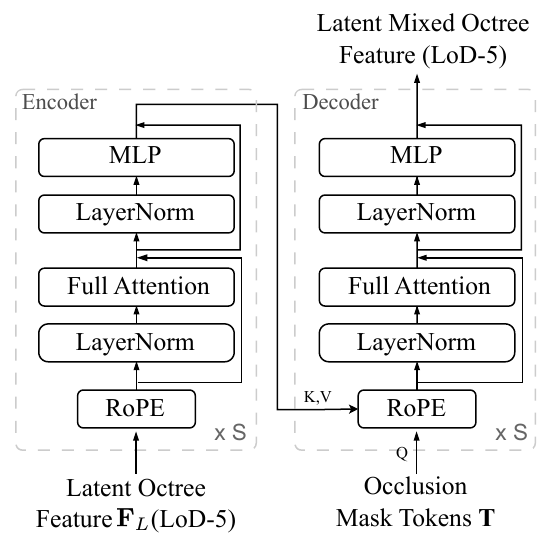}%
        \caption{Overall architecture of Latent 3D MAE.}
        \label{fig:latent_3d_mae}
    \end{minipage}
\end{figure}

\paragraph{\textbf{Occlusion masking}.}
Next, we concatenate mask tokens $\mathbf{T} \in \mathbb{R}^{M\times D^{\prime}}$ to the encoded latent octree feature where $M$ is the number of the mask tokens. Note that each of the mask tokens has identical learnable parameters.
The key question is how to place them in 3D space.
Although previous methods~\cite{li2023voxformer} put mask tokens inside all the empty cells of a dense voxel grid, it is unlikely that visible regions extending from the camera to the input depth are occupied unless the error of a depth map is enormous.
Further, this dense masking strategy forces to use a local attention mechanism such as deformable 3D attention used in VoxFormer~\cite{li2023voxformer}, due to the highly expensive memory and computational cost.
To address this issue, we introduce an occlusion masking strategy in which the mask tokens $\mathbf{T}$ are placed only into occluded voxels. Concretely, we perform depth testing on every voxel within a voxel grid to determine if they are positioned behind objects. Mask tokens are assigned to their respective locations only after passing this test. The proposed occlusion masking strategy and efficient positional encoding enable our latent 3D MAE (\Cref{fig:latent_3d_mae}) to leverage full attention instead of local attention.

\paragraph{\textbf{Decoder architecture}.}
The masked octree feature is given to the latent 3D MAE decoder which consists of $S$ layers of the full cross-attention Transformer blocks with RoPE~\cite{su2021roformer} to learn global reasoning including occluded regions. Finally, the decoder of Octree U-Net takes the mixed latent octree feature of the Transformer decoder $\mathbf{F}_{ML} \in \mathbb{R}^{\left(N^\prime + M \right) \times D^\prime}$ as input and upsamples features with skip connections. The decoded feature is passed to a two-layer MLP which estimates an occupancy at LoD-$h$. In addition, normals and SDF values are predicted only at the final LoD. 
To avoid unnecessary computation, we prune grid cells predicted as empty with a threshold of $0.5$ at every LoD, following~\cite{PSWang2020}. 

\subsection{Training Details and Loss Functions}
We use all surface points extracted through OpenVDB~\cite{10.1145/2487228.2487235} during training.
The loss function is defined as

\begin{equation}
  \mathcal{L} = \mathcal{L}_{nrm} + \mathcal{L}_{SDF} + \sum_{h \in \{5, 6, 7, 8, 9\}} \mathcal{L}^h_{occ},
\end{equation}
where $\mathcal{L}_{nrm}$ and $\mathcal{L}_{SDF}$ measure the averaged L2 norm of normals and SDF values. $\mathcal{L}^h_{occ}$ computes a mean of binary cross entropy function of each LoD-{h}.

\begin{table}[t!]
    \caption{\textbf{Dataset comparisons.} We create the first large-scale and diverse 3D scene completion dataset for novel multiple objects using a subset of 3D models from Objaverse dataset~\cite{objaverse}. The number of categories is reported by using the LVIS categories, and $R^{\text{LVIS}}$(\%) represents a ratio of the number of the categories covered by the dataset. $\dag$ denotes the number of objects with actual size. }
    \centering
    \setlength{\tabcolsep}{8pt}
    \scalebox{0.9}{
    \begin{tabular}{c|cccccc}
       \multirow{2}{*}{Dataset}  & \multirow{2}{*}{Type} & 3D & \# & \# & \# & \multirow{2}{*}{$R^{\text{LVIS}}$(\%)} \\
       & & Models & Frames & Objs & Cats \\ \hline
       YCB-V~\cite{xiang2018posecnn} & Real & \checkmark & $133$K & $21$ & $5$ & $0.4$ \\
       HB~\cite{kaskman2019homebreweddb} & Real & \checkmark & $17$K & $33$ & $13$ & $1.0$ \\
       HOPE~\cite{lin2021fusion} & Real & \checkmark & $2$K & $28$ & $3$ & $0.3$ \\
       CO3D V2~\cite{reizenstein21co3d} & Real & & $6$M & $\textbf{40}$\textbf{K} & \underline{$50$} & \underline{$4.2$} \\
       MegaPose~\cite{labbe2022megapose} & Synthetic & \checkmark & $1$M & $1\text{K}^\dag$ & 17 & 0.9 \\
       Ours & Synthetic & \checkmark & $1$M & $\underline{12}$\underline{K} & $\textbf{601}$ & $\textbf{50.0}$ \\
    \end{tabular}}
    \label{tab:dataset}
\end{table}

\section{Dataset}
\label{sec:dataset}
As shown in \Cref{tab:dataset}, existing datasets are limited in the diversity of object categories. Although the CO3D V2 dataset~\cite{reizenstein21co3d} contains data for 40k objects, because the provided ground-truth 3D shapes are reconstructed from unposed multi-view images, they tend to be highly noisy and parts of the object missing due to lack of visibility. To tackle this problem, we leverage Objaverse~\cite{objaverse}, a large-scale $1$M 3D object dataset containing $46$k objects with LVIS category annotations. To focus on completion of hand-held objects, we select $601$ categories and ensure that the largest dimension of the objects in each category falls approximately within the range of $4$ cm to $40$ cm. In addition, for high-quality rendering, we omit objects that lack textures, contain more than $10,000$ vertices, or are articulated. To increase the number of objects, we add objects from Google Scanned Objects (GSO)~\cite{10.1109/ICRA46639.2022.9811809}, which results in $12,655$ objects in total. We render $1$M images of $25,000$ scenes using physics-based rendering and positioning via BlenderProc~\cite{Denninger2023} to simulate realistic scenes (\Cref{fig:dataset_example}). For each image, we randomly choose a camera view such that at least one object is within the camera frame. We also generate $1,000$ images using $250$ withheld objects for evaluation.

\section{Experimental Results}
\label{sec:experimental_results}

\paragraph{\textbf{Implementation details.}}
We train all the models for $2
$ epochs using the Adam~\cite{kingma:adam} optimizer with a learning rate of $0.002$ and batch size of $16$ on NVIDIA A100. Note that the models are only trained on the synthetic dataset introduced in \Cref{sec:dataset}. In addition, the number of Transformer blocks $K$, the feature dimension $D$, and $D^\prime$ are set to $3$, $32$, and $192$ respectively. We use a pre-trained model of ResNeXt-50~\cite{Xie2016} as an image encoder for all the experiments. The ground-truth occupancy, SDF and normals are computed from meshes with OpenVDB~\cite{10.1145/2487228.2487235}. During training, we dilate ground-truth masks using the radius randomly selected from $1$, $3$ and $5$ pixels to deal with the segmentation error around the object edges. During evaluation, we use ground-truth masks provided by the datasets.

\paragraph{\textbf{Evaluation metrics.}}
We report Chamfer distance (CD), F1-Score@10mm (F1), and normal consistency (NC) to evaluate the quality of a completed surface. For surface-based methods, we use a predicted surface directly for evaluation. For the methods that predict occupancy, the marching cubes algorithm~\cite{10.1145/37402.37422} is used to extract a surface and uniformly sample $100,000$ points from its surface such that the number of points are roughly equal to the surface prediction methods. We use mm as a unit for all the reported metrics.

\paragraph{\textbf{Evaluation datasets.}}
We evaluate the baselines and our model on one synthetic and three real-world datasets.
For the synthetic dataset, we render $1,000$ images using textured 3D scans from Objaverse~\cite{objaverse}, following the same procedure described in \Cref{sec:dataset}. We randomly choose $3$ to $5$ objects per image from the withheld objects for Objavese dataset. Since these 3D scans are relatively more complex than the objects seen in the real-world datasets we use, they can provide a good scene completion quality estimate for complex objects. For the real-world dataset, we use the YCB-Video~\cite{xiang2018posecnn}, HOPE~\cite{lin2021fusion} and HomebrewedDB (HB)~\cite{kaskman2019homebreweddb} datasets. YCB-Video consists of $21$ everyday objects with diverse shapes. HOPE contains $28$ simple household objects with mostly rectangular and cylindrical everyday shapes, and the images are captured in various lighting conditions in indoor scenes using a RealSense D415 RGBD camera. HB includes $33$ objects (\eg, toy, household, and industrial objects). Their images are taken by PrimeSense Carmine in lab-like environments.

\paragraph{\textbf{Baselines.}}
As discussed in \cref{sec:intro,sec:related_work}, multi-object scene completion from a single RGB-D image is relatively not explored due to the lack of large-scale and diverse multi-object scene completion datasets. We carefully choose baseline architectures that can support this task with simple or no adaptation. We focus on three primary method types from related fields. Firstly, we select Semantic Scene Completion (SSC) methods~\cite{Li2020aicnet,PSWang2020,choy20194d,li2023voxformer} that do not heavily rely on domain or categorical knowledge of indoor or outdoor scenes. Secondly, we opt for object shape completion methods~\cite{choy20194d,wu2023multiview,PSWang2020,yan2022shapeformer} that can be extended to multi-object scene completion without an architectural modification and prohibitive memory utilization. Thirdly, we consider voxel or octree-based 3D reconstruction methods~\cite{Peng2020ECCV,PSWang2020,choy20194d,Boulch_2022_CVPR} that predict a complete and plausible shape using noisy and sparse point cloud data. For dense voxel-based (\eg, AICNet~\cite{Li2020aicnet}, ConvONet~\cite{Peng2020ECCV} and VoxFormer~\cite{li2023voxformer}) and sparse voxel-based methods (\eg, MinkowskiNet~\cite{choy20194d}, OCNN~\cite{PSWang2020}, and our method), we use LoD-$6$ and LoD-$9$ as an input resolution respectively. All the experiments are conducted using the original implementation provided by the authors, with few simple modifications to adapt for multi-object scene completion and a fair comparison. For instance, we extend the baselines that take the point cloud as input by concatenating the image features to the point cloud features. For occupancy-based methods, though their output voxel grid resolution is LoD-$6$, we use trilinear interpolation to predict occupancy at LoD-$7$~\cite{Peng2020ECCV}. For MinkowskiNet~\cite{choy20194d} and OCNN~\cite{10.1145/3072959.3073608,PSWang2020}, we use the U-Net architecture with the depth of $5$ (LoD-$9$ to LoD-$4$).
We discuss further details about the baseline architectures, their modifications, and hyperparameters in the supplemental.

\begin{table*}[t]
    \caption{Quantitative evaluation of multi-object scene completion on Ours, YCB-Video~\cite{xiang2018posecnn}, HOPE~\cite{lin2021fusion}, and HomebrewedDB~\cite{kaskman2019homebreweddb} datasets. Chamfer distance (CD), F1-Score@10mm (F1), and normal consistency (NC) are reported. Chamfer distance is reported in the unit of mm.}
    \centering
    \scalebox{0.8}{\begin{tabular}{l||c||ccc|ccc|ccc|ccc}
        \multirow{3}{*}{Method} & \multirow{3}{*}{3D Rep.} & \multicolumn{3}{c|}{Synthetic} & \multicolumn{9}{c}{Real} \\ \cline{3-14} \addlinespace[1pt]
        & & \multicolumn{3}{c|}{Ours} & \multicolumn{3}{c|}{YCB-Video~\cite{xiang2018posecnn}} & \multicolumn{3}{c|} {HB~\cite{kaskman2019homebreweddb}} &\multicolumn{3}{c}{HOPE~\cite{lin2021fusion}} \\ \cline{3-14} \addlinespace[1pt]
        & & CD$\downarrow$ & F1$\uparrow$ & NC$\uparrow$ & CD$\downarrow$ & F1$\uparrow$ & NC$\uparrow$ & CD$\downarrow$ & F1$\uparrow$ & NC$\uparrow$ & CD$\downarrow$ & F1$\uparrow$ & NC$\uparrow$  \\ \hline \addlinespace[1pt]
        VoxFormer~\cite{li2023voxformer} & Dense & 44.54 & 0.382 & 0.653 & 30.32 & 0.438 & 0.641 & 34.84 & 0.366 & 0.608 & 47.75 & 0.323 & 0.594 \\
        ShapeFormer~\cite{yan2022shapeformer} & Dense & 39.50 & 0.401 & 0.593 & 38.21 & 0.385 & 0.588 & 40.93 & 0.328 & 0.594 & 39.54 & 0.306 & 0.591 \\
        MCC~\cite{wu2023multiview} & Implicit & 43.37 & 0.459 & 0.700 & 35.85 & 0.289 & 0.608 & 19.59 & 0.371 & 0.655 & 17.53 & 0.357 & 0.658 \\
        ConvONet~\cite{Peng2020ECCV} & Dense & 23.68 & 0.541 & 0.710 & 32.87 & 0.458 & 0.649 & 26.71 & 0.504 & 0.643 & 20.95 & 0.581 & 0.678 \\
        POCO~\cite{Boulch_2022_CVPR} & Implicit & 21.11 & 0.634 & 0.753 & 15.45 & 0.587 & 0.699 & 13.17 & 0.624 & 0.709 & 13.20 & 0.602 & 0.706 \\
        AICNet~\cite{Li2020aicnet} & Dense & 15.64 & 0.573 & 0.741 & 12.26 & 0.545 & 0.702 & 11.87 & 0.557 & 0.674 & 11.40 & 0.564 & 0.670 \\
        Minkowski~\cite{choy20194d} & Sparse & 11.47 & 0.746 & 0.802 & 8.04 & 0.761 & 0.717 & 8.81 & 0.728 & 0.719 & 8.56 & 0.734 & 0.709 \\
        OCNN~\cite{PSWang2020} & Sparse & \underline{9.05} & \underline{0.782} & \underline{0.828} & \underline{7.10} & \underline{0.778} & \underline{0.771} & \underline{7.02} & \underline{0.792} & \underline{0.736} & \underline{8.05} & \underline{0.742} & \underline{0.736} \\\hline \addlinespace[2pt]
        Ours & Sparse & \textbf{6.48} & \textbf{0.839} & \textbf{0.848} & \textbf{6.40} & \textbf{0.800} & \textbf{0.785} & \textbf{6.14} & \textbf{0.819} & \textbf{0.770} & \textbf{6.97} & \textbf{0.803} & \textbf{0.750} \\
    \end{tabular}}
    \label{tab:main_result}
\end{table*}

\subsection{Quantitative Results}
\begin{figure*}[h!]
    \begin{minipage}{\textwidth}
        \centering
        \includegraphics[width=\textwidth]{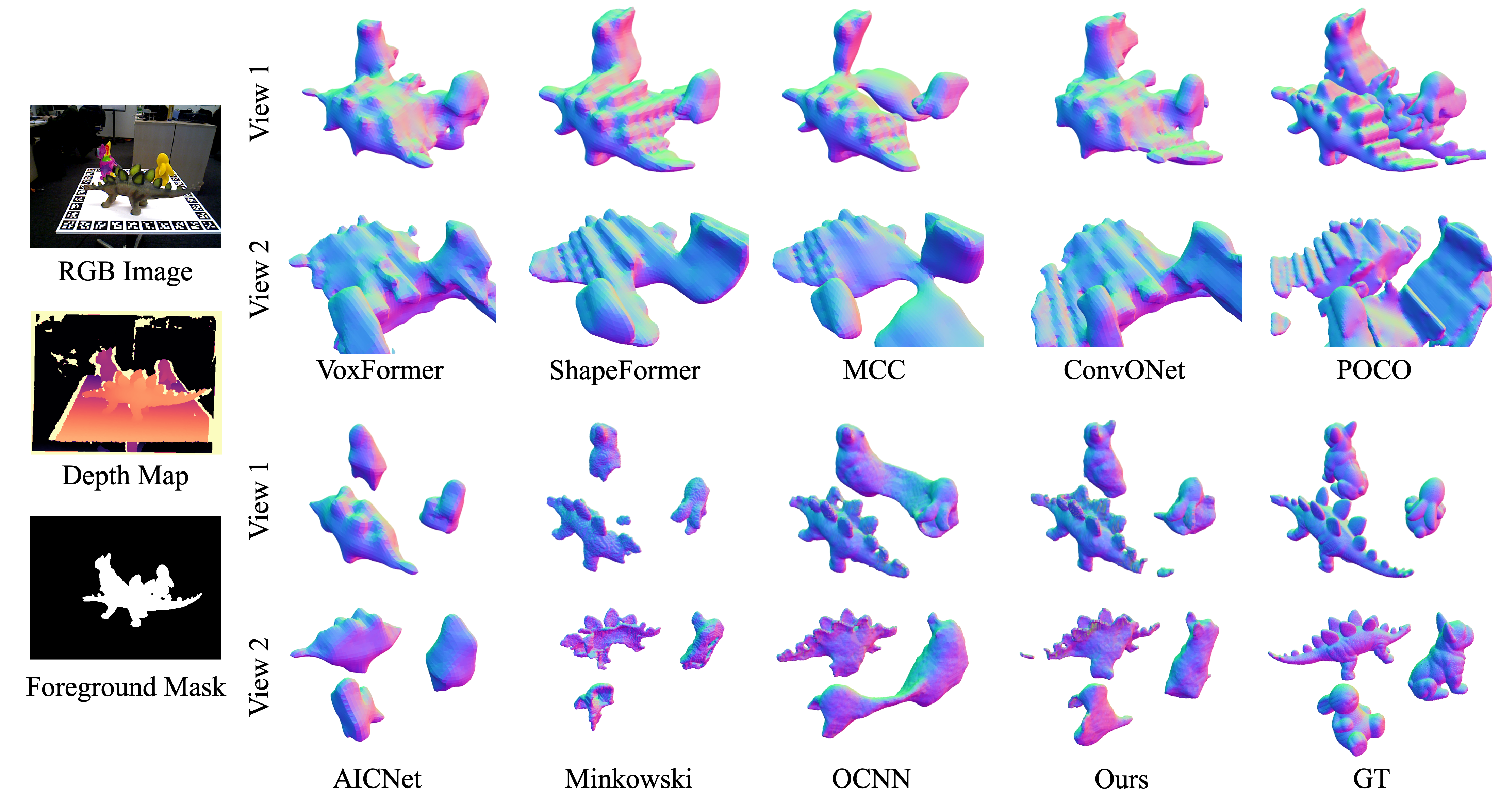}
    \end{minipage}
 \\
     \begin{minipage}{\textwidth}
        \centering
        \includegraphics[width=\textwidth]{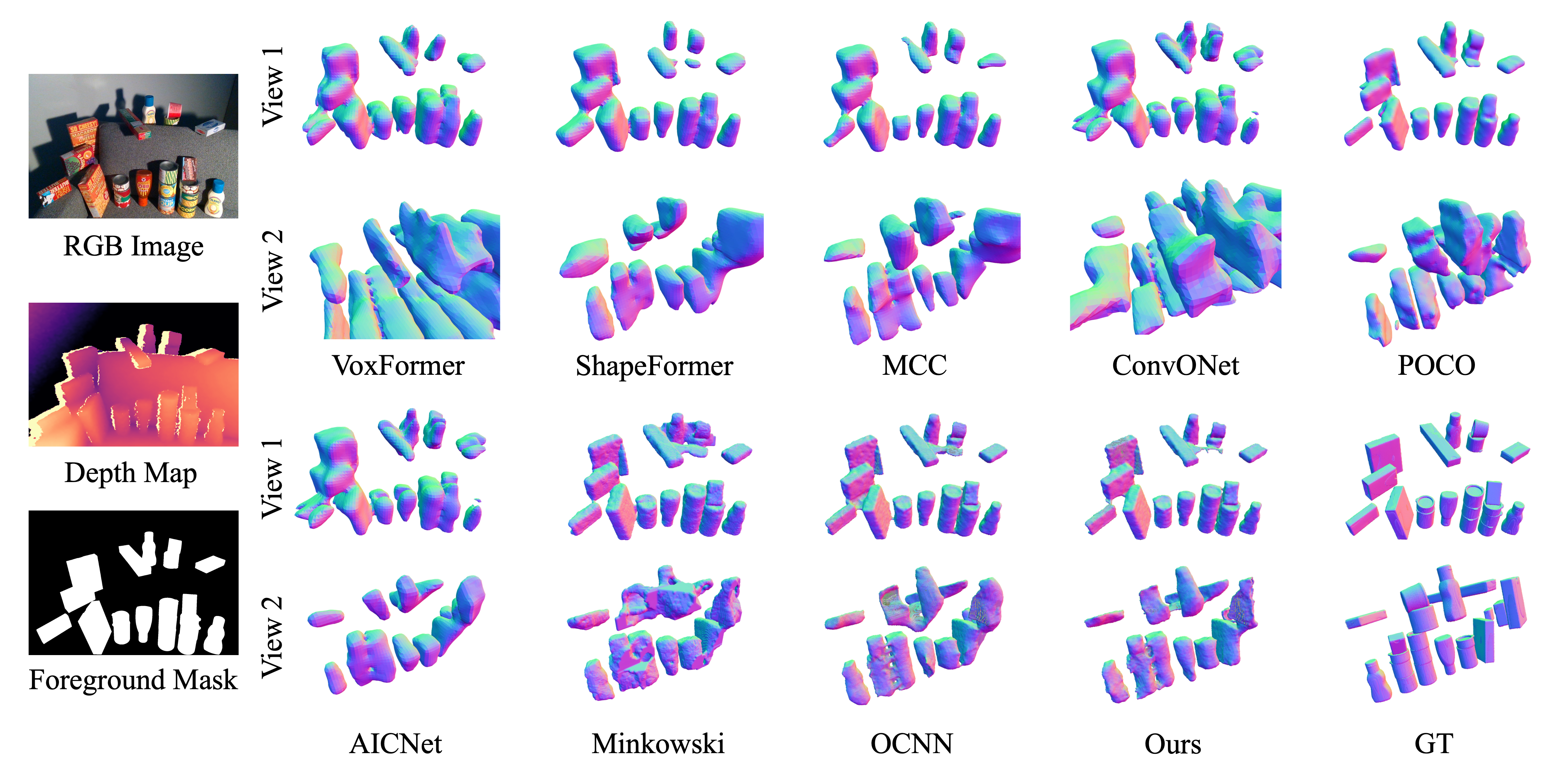}
    \end{minipage}
    \caption{Comparisons on HomebrewedDB dataset (\textbf{Top}), and HOPE (\textbf{Bottom}) datasets. For better visibility, we show the generated and ground truth shapes. The top and bottom rows show an image from near camera and back views respectively. Compared to the other methods, our method predicts accurate and consistent shapes on a challenging scene completion task for novel objects.}
    \label{fig:experimental_resutls}
\end{figure*}

\Cref{tab:main_result} shows that our method outperforms the baselines on all the metrics and datasets. Although our model is only trained on synthetic data, it demonstrates strong generalizability to real-world datasets.
We also remark that our method exhibits robustness to the noise characteristics present in depth data captured by typical RGB-D cameras despite being trained on noise-free depth data in simulation. The comparisons show that hierarchical structures and the latent 3D MAE are key to predicting 3D shapes of unseen objects more accurately than the baselines. Unlike our method, VoxFormer~\cite{li2023voxformer} uses an MAE with 3D deformable attention where only $8$ neighbors of the reference points at the finest resolution are considered.
\Cref{fig:experimental_resutls} also demonstrates that methods using a dense voxel grid or implicit representation fail to generalize to novel shapes. This implies that capturing a right choice of a network architecture is crucial to learn generalizable shape priors for zero-shot multi-object scene completion.
Our method has the similar U-Net architecture used in MinkowskiNet~\cite{choy20194d} and OCNN~\cite{10.1145/3072959.3073608} except we use the latent 3D MAE at LoD-$5$ instead of making the network deeper. This indicates that the latent 3D MAE can better approximate the shape distribution of the training dataset by leveraging an attention mechanism to capture global 3D contexts. \Cref{tab:runtime} also confirms that our method achieves the best scene completion quality by measuring Chamfer distance in visible and occluded regions separately.

\begin{table}[t!]
\centering
\begin{minipage}{.40\linewidth}
    \setlength{\tabcolsep}{5pt}
    \caption{Ablation Study of positional encoding on our synthetic dataset. We compare w/o positional encoding, conditional positional encoding (CPE)~\cite{chu2023CPVT}, absolute postional encoding (APE) used in \cite{li2023voxformer}, and RoPE~\cite{su2021roformer}. }
    \centering
    \scalebox{0.9}{
    \begin{tabular}{l|ccc}
       Type  & CD$\downarrow$ & F1$\uparrow$ & NC$\uparrow$ \\ \hline \addlinespace[2pt]
       w/o  & 11.32 & 0.778 & 0.808 \\
       CPE~\cite{chu2023CPVT}  & 9.91 & 0.785 & 0.811 \\
       APE~\cite{li2023voxformer}  & 8.61 & 0.782 & 0.825 \\
       RPE~\cite{Wang2023OctFormer}  & \underline{7.81} & \underline{0.804} & \underline{0.830} \\
       RoPE~\cite{su2021roformer}  & \textbf{6.48} & \textbf{0.839} & \textbf{0.848} \\
    \end{tabular}}
    \label{tab:positional_encoding}
\end{minipage} \quad
\begin{minipage}{.56\linewidth}
    \centering
    \caption{Ablation study on 3D attention algorithms. The scores are reported on the HOPE dataset~\cite{lin2021fusion}.}
    \scalebox{0.9}{
    \begin{tabular}{l|c|ccc}
       \multirow{2}{*}{Method} & Occ. & \multirow{2}{*}{CD$\downarrow$} & \multirow{2}{*}{F1$\uparrow$} & \multirow{2}{*}{Runtime$\downarrow$} \\
       & Masking & & & \\ \hline \addlinespace[2pt]
       3D DSA~\cite{li2023voxformer} & & 12.14 & 0.703 & 93.3 \\
       Neighbor. Attn.~\cite{zhao2021point} & & 9.26 & 0.727 & 130.8 \\
       Octree Attn.~\cite{Wang2023OctFormer} & & 7.99 & 0.752 & 116.4 \\ \hline
       Neighbor. Attn.~\cite{zhao2021point} & \checkmark & 8.81 & 0.759 & 111.9 \\
       Octree Attn.~\cite{Wang2023OctFormer} & \checkmark & 7.54 & 0.772 & 105.3 \\
       Full + Self Attn. & \checkmark & \underline{7.21} & \underline{0.785} & \underline{86.2}  \\
       Full + Cross Attn. & \checkmark & \textbf{6.97} & \textbf{0.803} & \textbf{85.1} \\
    \end{tabular}}
    \label{tab:attention_alg}
\end{minipage}
\end{table}

\paragraph{\textbf{Positional encoding.}}
As shown in \Cref{tab:positional_encoding}, we explore the effect of RoPE~\cite{su2021roformer} on the validation set of our synthetic dataset. The first row shows that all the metrics significantly drop if positional encoding is not used. In addition, we test CPE~\cite{chu2023CPVT}, APE~\cite{li2023voxformer}, and RPE~\cite{Wang2023OctFormer} and obtain slightly better scores. CPE~\cite{chu2023CPVT} is typically more effective than APE in tasks such as 3D instance/semantic segmentation and object detection where a complete 3D point cloud is given. However, this result highlights the challenge of capturing position information from mask tokens which initially have the identical parameters. Our method employs RoPE~\cite{su2021roformer} for relative positional embedding. One of the important aspect of RoPE~\cite{su2021roformer} is that it does not have any learnable parameters. Despite this, it demonstrates superior performance compared to other approaches. Although RoPE was originally proposed in the domain of natural language processing, our experiment reveals its effectiveness in multi-object 3D scene completion.

\paragraph{\textbf{3D Attention algorithms.}}
 \Cref{tab:attention_alg} reveals that occlusion masking yields better runtime and metrics than dense masking. Furthermore, our experiments suggest that full attention and Octree attention, both characterized by their wider receptive fields, are more effective compared to local attention algorithms such as 3D deformable self-attention (3D DSA)~\cite{li2023voxformer} and neighborhood attention~\cite{zhao2021point}.

\begin{table}[t]
\begin{minipage}{.40\linewidth}
    \centering
    \caption{Ablation study of the number of MAE layers on our synthetic dataset.}
    \scalebox{0.9}{
    \begin{tabular}{c|cccc}
       \#Layers  & CD$\downarrow$ & F1$\uparrow$ & NC$\uparrow$ & Runtime$\downarrow$ \\ \hline\addlinespace[2pt]
       1 & 9.01 & 0.784 & 0.828 & \textbf{76.4} \\
       3  & \underline{6.48} & \underline{0.839} & \underline{0.848} & \underline{85.1} \\
       5  & \textbf{5.75} & \textbf{0.850} & \textbf{0.855} & 96.2 \\
    \end{tabular}}
    \label{tab:num_layers}
\end{minipage} \quad
\begin{minipage}{.56\linewidth}
    \caption{Ablation study of U-Net architectures on HomebrewedDB dataset~\cite{kaskman2019homebreweddb}.}
    \centering
    \scalebox{0.9}{
    \begin{tabular}{l|cccc}
       Architecture  & CD$\downarrow$ & F1$\uparrow$ & NC$\uparrow$ & Runtime$\downarrow$ \\ \hline \addlinespace[2pt]
       Mink. U-Net~\cite{choy20194d}  & \underline{7.26} & \underline{0.788} & \underline{0.743} & \textbf{83.8} \\
       OctFormer~\cite{Wang2023OctFormer}  &  7.45 & 0.756 & 0.728 & 114.4 \\
       Octree U-Net~\cite{10.1145/3072959.3073608}  & \textbf{6.14} & \textbf{0.819} & \textbf{0.770} & \underline{85.1} \\
    \end{tabular}}
    \label{tab:unet_layers}
\end{minipage}
\end{table}

\begin{table}[t]
    \setlength{\tabcolsep}{5pt}
    \caption{Comparisons of the runtime (ms). For reference, we also show Chamfer distance of visible $\text{CD}_{vis}$ and occluded $\text{CD}_{occ}$ regions on our synthetic dataset.}
    \centering
    \scalebox{1.0}{
    \begin{tabular}{l|cc|ccc|c}
       Method & 3D Rep. & Resolution & $\text{CD}_{vis}$ $\downarrow$ & $\text{CD}_{occ}$$\downarrow$ & CD$\downarrow$ & Runtime$\downarrow$ \\ \hline
       VoxFormer~\cite{li2023voxformer} & Dense & $128^3$ & $18.25$ & $66.32$ & $44.54$ & $79.5$ \\ 
       ShapeFormer~\cite{yan2022shapeformer} & Dense & $128^3$ & $14.61$ & $63.33$ & $39.50$ & $1.8 \times 10^4$ \\
       MCC~\cite{wu2023multiview} & Implicit & $128^3$ & $15.39$ & $63.41$ & $44.37$ & $9.1 \times 10^3$ \\ 
       ConvONet~\cite{Peng2020ECCV} & Dense & $128^3$ & $17.09$ & $34.09$ & $23.68$ & \underline{48.4} \\
       POCO~\cite{Boulch_2022_CVPR} & Implicit & $128^3$ & $10.37$ & $31.55$ & $21.11$ & $758.8$ \\
       AICNet~\cite{Li2020aicnet} & Dense & $128^3$ & $9.98$ & $21.43$ & $15.64$ & \textbf{24.2} \\ \hline \addlinespace[2pt]
       Minkowski~\cite{choy20194d} & Sparse & $512^3$ & $7.12$ & $15.44$ & $11.47$ & $78.5$ \\ 
       OCNN~\cite{PSWang2020} & Sparse & $512^3$ & $\underline{3.87}$ & $\underline{12.16}$ & $\underline{9.05}$  & $80.1$ \\ \hline \addlinespace[2pt]
       Ours & Sparse & $512^3$ & $\textbf{3.29}$ & $\textbf{9.40}$ & $\textbf{6.48}$ & $85.1$ \\
    \end{tabular}}
    \label{tab:runtime}
\end{table}

\paragraph{\textbf{Number of layers in 3D latent MAE.}}
We further explore the design of 3D latent MAE in \Cref{tab:num_layers}. Increasing the number of layers in 3D latent MAE improves the scene completion quality while making the runtime slower. Consequently, we select $3$ layers for a good trade-off between the accuracy and runtime.

\paragraph{\textbf{U-Net architectures.}}
In \Cref{tab:unet_layers}, we investigate U-Net architectures. The key difference of Minkowski U-Net~\cite{choy20194d} is the use of a sparse tensor as an underlying data structure instead of an octree, which gives a slightly better performance than Octree U-Net~\cite{10.1145/3072959.3073608}. OctFormer~\cite{Wang2023OctFormer} proposes an octree-based window attention mechanism using the 3D Z-order curve to support a much larger kernel size than Octree U-Net. In general, a wider range of an effective receptive field helps achieve better performance. Nonetheless, OctFormer achieves a chamfer distance and F-1 score of $7.45$ and $0.756$, which is worse than Octree U-Net by $1.31$ and $0.063$ respectively. This indicates that the OctFormer's attention mechanism is less effective compared to an Octree U-Net architecture especially in the presence of latent 3D MAE, playing the similar role in the latent space.

\paragraph{\textbf{Runtime analysis.}}
\Cref{tab:runtime} shows the runtime performance of the baselines and our method. For a fair comparison, we run inference over the $50$ samples of the HOPE dataset and report the average time. For occupancy-based methods, we predict occupancy on object surfaces and occluded regions. Due to the memory-intensive nature of  MCC~\cite{Boulch_2022_CVPR}'s Transformer architecture, we run inference multiple times with the maximum chunk size of $10,000$ points.
Our experiments demonstrate that implicit 3D representations used in POCO~\cite{Boulch_2022_CVPR} and MCC~\cite{wu2023multiview} become slower when the voxel grid resolution is higher. Further, an autoregressive Transformer adopted in ShapeFormer~\cite{yan2022shapeformer} greatly increases the runtime.
Conversely, the methods which leverage sparse voxel grids (\eg, MinkowskiNet~\cite{choy20194d}, OCNN~\cite{PSWang2020}, and Ours) achieve much faster runtime thanks to efficient sparse 3D convolutions, and hierarchical pruning on predicted surfaces. Our method offers runtimes comparable to the fastest method, while implementing attention operations over the scene via latent 3D MAE, and achieving superior reconstruction.

\begin{figure}[t]
    \centering
    \begin{minipage}{0.48\textwidth}
     \includegraphics[width=\textwidth]{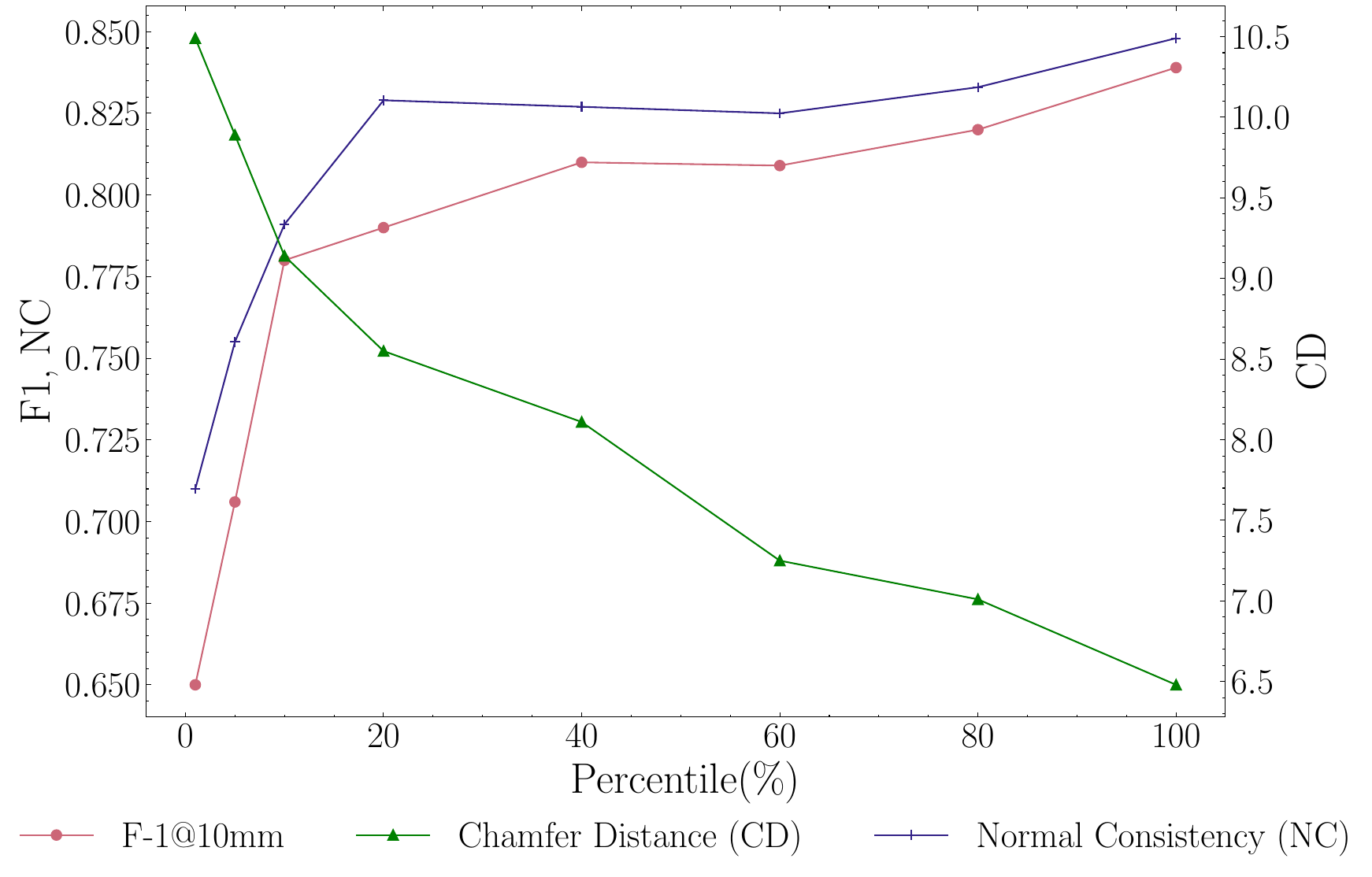}
    \label{fig:dataset_vs_performance}
    \caption{Scaling of the metrics with the number of objects in a training dataset. We conduct the experiments by changing the ratio of the number of objects to $1$\%, $5$\%, $10$\%, $20$\%, $40$\%, $60$\%, $80$\%, and $100$\%. }       
    \end{minipage} \quad
    \begin{minipage}{0.48\textwidth}
    \centering
    \includegraphics[width=0.95\textwidth]{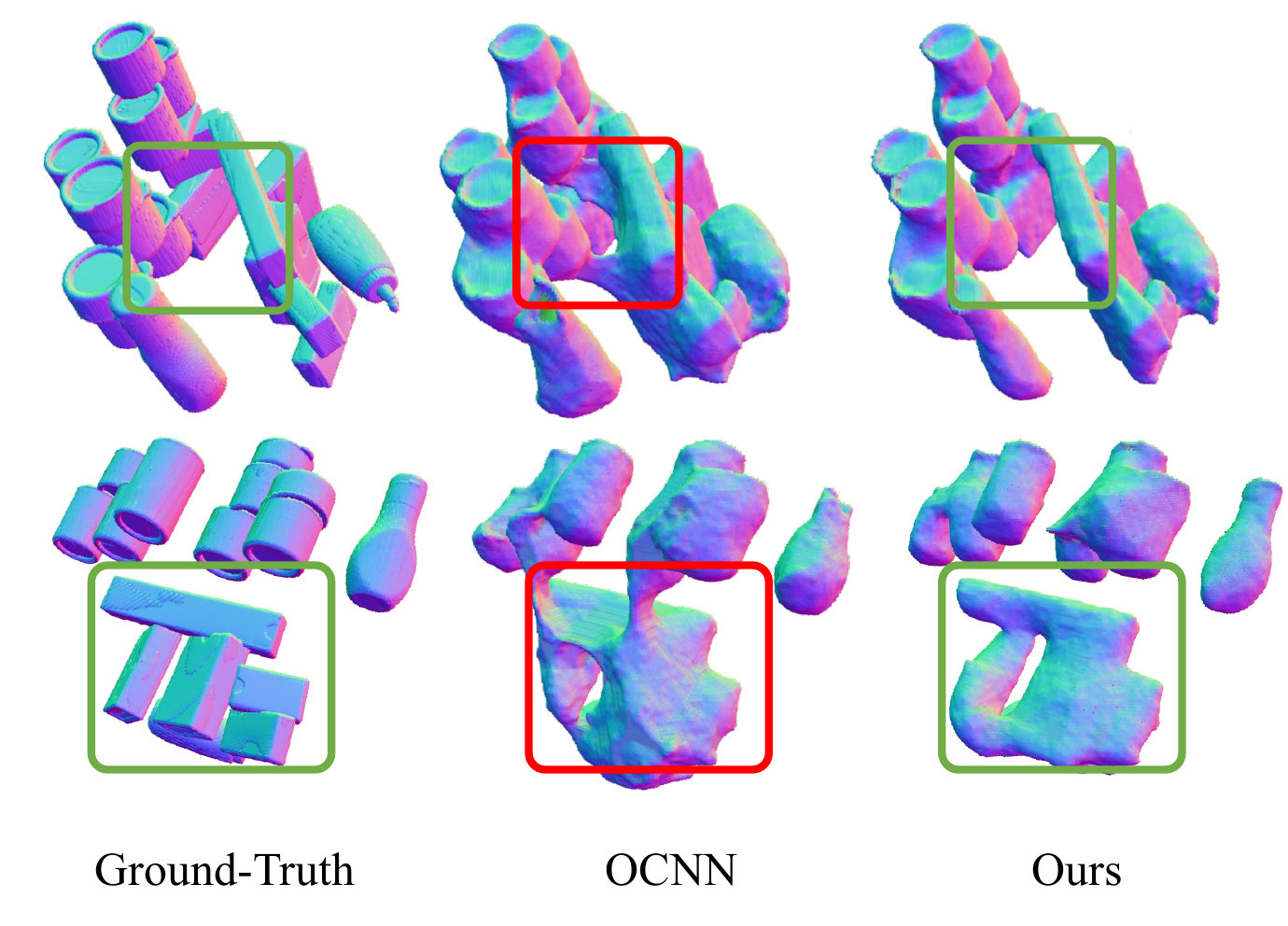}
    \caption{Qualitative comparison of OCNN~\cite{10.1145/3072959.3073608} and our method. Our proposed latent 3D MAE helps predict globally consistent scene completion.}       
    \label{fig:ablation_ocnn}
    \end{minipage}
\end{figure}

\paragraph{\textbf{Dataset scale analysis.}}
To assess the importance of the large-scale 3D scene completion datasets, we train our model on splits of increasing sizes which contain $1$\%, $5$\%, $10$\%, $20$\%, $40$\%, $60$\%, $80$\%, and $100$\% of the total number of the objects in our dataset. We report metrics on the test split of our dataset.
\Cref{fig:dataset_vs_performance} shows that all the metrics have a strong correlation with respect to the number of objects. This could imply that the model benefits significantly from increased data diversity and volume, enhancing its ability to understand and complete 3D shapes. We believe that this analysis is crucial for understanding the relationship between data quantity and model performance.

\begin{figure*}[t!]
    \centering
    \includegraphics[width=\textwidth]{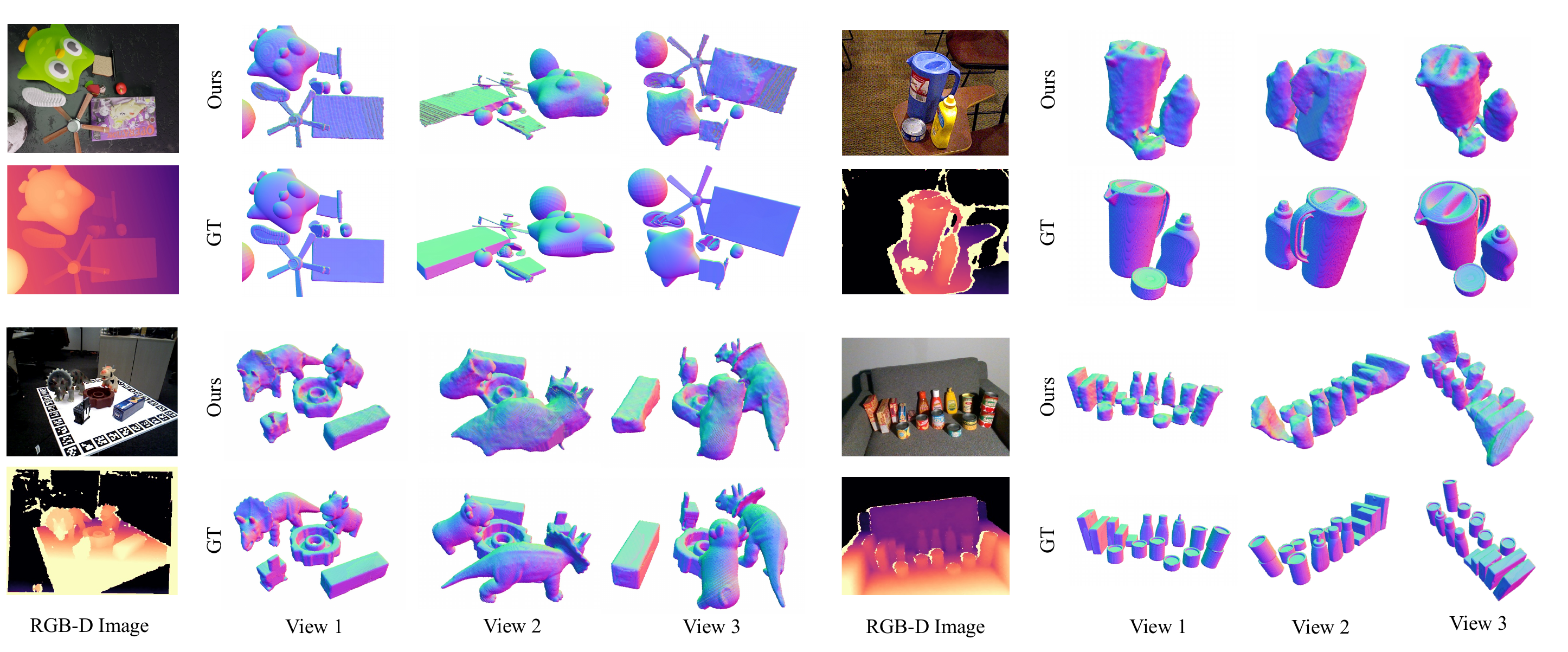}
    \caption{Qualitative results on our synthetic dataset (\textbf{Top Left}), YCB-Video (\textbf{Top Right}), HomebrewedDB (\textbf{Bottom Left}), and HOPE (\textbf{Bottom Right}) datasets. These results demonstrate the strong generalization to the real-world images on multi-object scene completion. We choose $3$ different views for better visibility. }
    \label{fig:qualitative_results}
\end{figure*}

\subsection{Qualitative Results}
\Cref{fig:qualitative_results} shows the qualitative results of our method on both of the synthetic and real-world datasets from three different views. Unlike the synthetic dataset, the real-world depth measurements are more noisy and erroneous, however, we observe that our method can generate faithful and consistent 3D shapes on different types of objects. These results indicate that our model successfully learns geometric and semantic priors of real-world objects only from the synthetic data. Moreover, \Cref{fig:ablation_ocnn} provides a comparison between our method and the second-best baseline, OCNN~\cite{10.1145/3072959.3073608}. OCNN struggles with multi-object reasoning, resulting in unnatural artifacts improperly merging multiple objects. We believe this finding further supports that using 3D latent MAE helps capture a global context for better scene completion.

\section{Conclusion and Future Work}
\label{sec:discussion_and_future_work}
In this paper, we present OctMAE, a hybrid architecture combining an Octree U-Net and a latent 3D MAE, for efficient and generalizable scene completion. Further, we create the first large-scale and diverse 3D scene completion dataset, which consists of $1$M images rendered with $12$K objects with realistic scale. Our experimental results on a wide range of the datasets demonstrate accurate zero-shot multi-object scene completion is possible with a proper choice of the network architecture and dataset, which potentially facilitates several challenging robotics tasks such as robotic manipulation and motion planning. Although our method achieves superior performance, it comes with some limitations. First, truncated objects are not reconstructed properly since depth measurements are not available. We believe we can overcome this problem by incorporating techniques for query proposal~\cite{yu2021pointr} and amodal segmentation~\cite{zhu2017semantic}. The second limitation is that the semantic information of completed shapes is not predicted. Although our focus in this work is geometric scene completion, we believe it is an interesting direction to integrate a technique from an open-vocabulary segmentation methods to obtain instance-level completed shapes. Third, our method does not handle uncertainty of surface prediction explicitly. In future work, we plan to extend our method to model uncertainty to improve the scene completion quality and diversity.

\section*{Acknowledgment}
We thank Zubair Irshad and Jenny Nan for valuable feedback and comments. This research is supported by Toyota Research Institute.

\bibliographystyle{splncs04}
\bibliography{main}

\begin{thebibliography}{10}
\providecommand{\url}[1]{\texttt{#1}}
\providecommand{\urlprefix}{URL }
\providecommand{\doi}[1]{https://doi.org/#1}

\bibitem{Boulch_2022_CVPR}
Boulch, A., Marlet, R.: {POCO: Point Convolution for Surface Reconstruction}. In: CVPR (2022)

\bibitem{bozic2021transformerfusion}
Bozic, A., Palafox, P., Thies, J., Dai, A., Nie{\ss}ner, M.: {TransformerFusion}: Monocular rgb scene reconstruction using transformers. In: NeurIPS (2021)

\bibitem{chan2023genvs}
Chan, E.R., Nagano, K., Chan, M.A., Bergman, A.W., Park, J.J., Levy, A., Aittala, M., Mello, S.D., Karras, T., Wetzstein, G.: {GeNVS}: Generative novel view synthesis with {3D}-aware diffusion models. In: CoRR (2023)

\bibitem{10.1007/978-3-031-19824-3_30}
Chen, H.X., Huang, J., Mu, T.J., Hu, S.M.: {CIRCLE: Convolutional Implicit Reconstruction And Completion For Large-Scale Indoor Scene}. In: ECCV (2022)

\bibitem{cheng2023sdfusion}
Cheng, Y.C., Lee, H.Y., Tulyakov, S., Schwing, A.G., Gui, L.Y.: {SDFusion}: Multimodal 3d shape completion, reconstruction, and generation. In: CVPR (2023)

\bibitem{choy20194d}
Choy, C., Gwak, J., Savarese, S.: {4D Spatio-Temporal ConvNets: Minkowski Convolutional Neural Networks}. In: CVPR (2019)

\bibitem{chu2023CPVT}
Chu, X., Tian, Z., Zhang, B., Wang, X., Shen, C.: {Conditional Positional Encodings for Vision Transformers}. In: ICLR (2023)

\bibitem{together2023redpajama}
Computer, T.: {RedPajama: an Open Dataset for Training Large Language Models} (2023)

\bibitem{dai2020sgnn}
Dai, A., Diller, C., Nie{\ss}ner, M.: {SG-NN}: Sparse generative neural networks for self-supervised scene completion of rgb-d scans. In: CVPR (2020)

\bibitem{dai2018scancomplete}
Dai, A., Ritchie, D., Bokeloh, M., Reed, S., Sturm, J., Nie{\ss}ner, M.: {ScanComplete: Large-Scale Scene Completion and Semantic Segmentation for 3D Scans}. In: CVPR (2018)

\bibitem{dao2023flashattention2}
Dao, T.: Flash{A}ttention-2: Faster attention with better parallelism and work partitioning  (2023)

\bibitem{objaverse}
Deitke, M., Schwenk, D., Salvador, J., Weihs, L., Michel, O., VanderBilt, E., Schmidt, L., Ehsani, K., Kembhavi, A., Farhadi, A.: {Objaverse: A Universe of Annotated 3D Objects}. CVPR  (2022)

\bibitem{Denninger2023}
Denninger, M., Winkelbauer, D., Sundermeyer, M., Boerdijk, W., Knauer, M., Strobl, K.H., Humt, M., Triebel, R.: {BlenderProc2: A Procedural Pipeline for Photorealistic Rendering}. Journal of Open Source Software  (2023)

\bibitem{Devlin2019BERTPO}
Devlin, J., Chang, M.W., Lee, K., Toutanova, K.: {BERT: Pre-training of Deep Bidirectional Transformers for Language Understanding}. In: NAACL (2019)

\bibitem{dosovitskiy2020vit}
Dosovitskiy, A., Beyer, L., Kolesnikov, A., Weissenborn, D., Zhai, X., Unterthiner, T., Dehghani, M., Minderer, M., Heigold, G., Gelly, S., Uszkoreit, J., Houlsby, N.: {An Image is Worth 16x16 Words: Transformers for Image Recognition at Scale}. ICLR  (2021)

\bibitem{10.1109/ICRA46639.2022.9811809}
Downs, L., Francis, A., Koenig, N., Kinman, B., Hickman, R., Reymann, K., McHugh, T.B., Vanhoucke, V.: {Google Scanned Objects: A High-Quality Dataset of 3D Scanned Household Items}. In: ICRA (2022)

\bibitem{duan2020curriculum}
Duan, Y., Zhu, H., Wang, H., Yi, L., Nevatia, R., Guibas, L.J.: {Curriculum deepsdf}. In: ECCV (2020)

\bibitem{functa22}
Dupont, E., Kim, H., Eslami, S.M.A., Rezende, D.J., Rosenbaum, D.: {From data to functa: Your data point is a function and you can treat it like one}. In: ICML (2022)

\bibitem{gao2022convmae}
Gao, P., Ma, T., Li, H., Dai, J., Qiao, Y.: {ConvMAE: Masked Convolution Meets Masked Autoencoders}. NeurIPS  (2022)

\bibitem{goldblum2023no}
Goldblum, M., Finzi, M., Rowan, K., Wilson, A.G.: {The No Free Lunch Theorem, Kolmogorov Complexity, and the Role of Inductive Biases in Machine Learning}. CoRR  (2023)

\bibitem{3DSemanticSegmentationWithSubmanifoldSparseConvNet}
Graham, B., Engelcke, M., van~der Maaten, L.: {3D Semantic Segmentation with Submanifold Sparse Convolutional Networks}. CVPR  (2018)

\bibitem{he2022masked}
He, K., Chen, X., Xie, S., Li, Y., Doll{\'a}r, P., Girshick, R.: {Masked autoencoders are scalable vision learners}. In: CVPR (2022)

\bibitem{hou2020revealnet}
Hou, J., Dai, A., Nie{\ss}ner, M.: {RevealNet: Seeing Behind Objects in RGB-D Scans}. In: CVPR (2020)

\bibitem{huang2023nksr}
Huang, J., Gojcic, Z., Atzmon, M., Litany, O., Fidler, S., Williams, F.: {Neural Kernel Surface Reconstruction}. In: CVPR (2023)

\bibitem{huang2023zeroshape}
Huang, Z., Stojanov, S., Thai, A., Jampani, V., Rehg, J.M.: {ZeroShape: Regression-based Zero-shot Shape Reconstruction}. CVPR  (2023)

\bibitem{irshad2022shapo}
Irshad, M.Z., Zakharov, S., Ambrus, R., Kollar, T., Kira, Z., Gaidon, A.: {Shapo: Implicit representations for multi-object shape, appearance, and pose optimization}. In: ECCV (2022)

\bibitem{2017_rss_system}
Kappler, D., Meier, F., Issac, J., Mainprice, J., Garcia~Cifuentes, C., W{\"u}thrich, M., Berenz, V., Schaal, S., Ratliff, N., Bohg, J.: {Real-time Perception meets Reactive Motion Generation}. RA-L  (2018)

\bibitem{10.1177/0278364911406761}
Karaman, S., Frazzoli, E.: {Sampling-Based Algorithms for Optimal Motion Planning}. Int. J. Rob. Res.  (2011)

\bibitem{kaskman2019homebreweddb}
Kaskman, R., Zakharov, S., Shugurov, I., Ilic, S.: {HomebrewedDB: RGB-D Dataset for 6D Pose Estimation of 3D Objects}. ICCVW  (2019)

\bibitem{kim2022revisiting}
Kim, T., Kim, K., Lee, J., Cha, D., Lee, J., Kim, D.: {Revisiting Image Pyramid Structure for High Resolution Salient Object Detection}. In: Proceedings of the Asian Conference on Computer Vision. pp. 108--124 (2022)

\bibitem{kingma:adam}
Kingma, D.P., Ba, J.: {Adam: A method for stochastic optimization}. In: ICLR (2015)

\bibitem{labbe2022megapose}
Labb\'e, Y., Manuelli, L., Mousavian, A., Tyree, S., Birchfield, S., Tremblay, J., Carpentier, J., Aubry, M., Fox, D., Sivic, J.: {MegaPose}: 6d pose estimation of novel objects via render \& compare. In: CoRL (2022)

\bibitem{Li2020aicnet}
Li, J., Han, K., Wang, P., Liu, Y., Yuan, X.: {Anisotropic Convolutional Networks for 3D Semantic Scene Completion}. In: CVPR (2020)

\bibitem{Li2019ddr}
Li, J., Liu, Y., Gong, D., Shi, Q., Yuan, X., Zhao, C., Reid, I.: {RGBD Based Dimensional Decomposition Residual Network for 3D Semantic Scene Completion}. In: CVPR. pp. 7693--7702 (June 2019)

\bibitem{li2021grounded}
Li*, L.H., Zhang*, P., Zhang*, H., Yang, J., Li, C., Zhong, Y., Wang, L., Yuan, L., Zhang, L., Hwang, J.N., Chang, K.W., Gao, J.: Grounded language-image pre-training. In: CVPR (2022)

\bibitem{li2023voxformer}
Li, Y., Yu, Z., Choy, C., Xiao, C., Alvarez, J.M., Fidler, S., Feng, C., Anandkumar, A.: {VoxFormer: Sparse Voxel Transformer for Camera-based 3D Semantic Scene Completion}. In: CVPR (2023)

\bibitem{liang2023open}
Liang, F., Wu, B., Dai, X., Li, K., Zhao, Y., Zhang, H., Zhang, P., Vajda, P., Marculescu, D.: {Open-vocabulary semantic segmentation with mask-adapted clip}. In: CVPR (2023)

\bibitem{lin2021fusion}
Lin, Y., Tremblay, J., Tyree, S., Vela, P.A., Birchfield, S.: {Multi-view Fusion for Multi-level Robotic Scene Understanding}. In: IROS (2021)

\bibitem{liu2020neural}
Liu, L., Gu, J., Lin, K.Z., Chua, T.S., Theobalt, C.: {Neural Sparse Voxel Fields}. NeurIPS  (2020)

\bibitem{liu2023one}
Liu, M., Xu, C., Jin, H., Chen, L., Xu, Z., Su, H., et~al.: {One-2-3-45: Any single image to 3d mesh in 45 seconds without per-shape optimization}. NeurIPS  (2023)

\bibitem{liu2023zero1to3}
Liu, R., Wu, R., Hoorick, B.V., Tokmakov, P., Zakharov, S., Vondrick, C.: {Zero-1-to-3: Zero-shot One Image to 3D Object}. In: CVPR (2023)

\bibitem{Liu2023MeshDiffusion}
Liu, Z., Feng, Y., Black, M.J., Nowrouzezahrai, D., Paull, L., Liu, W.: {MeshDiffusion: Score-based Generative 3D Mesh Modeling}. In: ICLR (2023)

\bibitem{10.1145/37402.37422}
Lorensen, W.E., Cline, H.E.: {Marching Cubes: A High Resolution 3D Surface Construction Algorithm}. SIGGRAPH  (1987)

\bibitem{OccupancyNetworks}
Mescheder, L., Oechsle, M., Niemeyer, M., Nowozin, S., Geiger, A.: {Occupancy Networks: Learning 3D Reconstruction in Function Space}. In: CVPR (2019)

\bibitem{autosdf2022}
Mittal, P., Cheng, Y.C., Singh, M., Tulsiani, S.: {AutoSDF: Shape Priors for 3D Completion, Reconstruction and Generation}. In: CVPR (2022)

\bibitem{10160350}
Mohammadi, S.S., Duarte, N.F., Dimou, D., Wang, Y., Taiana, M., Morerio, P., Dehban, A., Moreno, P., Bernardino, A., Del~Bue, A., Santos-Victor, J.: {3DSGrasp: 3D Shape-Completion for Robotic Grasp}. In: ICRA (2023)

\bibitem{10.1145/2487228.2487235}
Museth, K.: {VDB: High-resolution sparse volumes with dynamic topology}  (2013)

\bibitem{okumura2023quick}
Okumura, K., Défago, X.: {Quick Multi-Robot Motion Planning by Combining Sampling and Search}. In: IJCAI (2023)

\bibitem{Park_2019_CVPR}
Park, J.J., Florence, P., Straub, J., Newcombe, R., Lovegrove, S.: {DeepSDF: Learning Continuous Signed Distance Functions for Shape Representation}. In: CVPR (2019)

\bibitem{Peng2020ECCV}
Peng, S., Niemeyer, M., Mescheder, L., Pollefeys, M., Geiger, A.: {Convolutional Occupancy Networks}. In: ECCV (2020)

\bibitem{rabe2021selfattention}
Rabe, M.N., Staats, C.: {Self-attention Does Not Need $O(n^2)$ Memory} (2021)

\bibitem{radford2021learning}
Radford, A., Kim, J.W., Hallacy, C., Ramesh, A., Goh, G., Agarwal, S., Sastry, G., Askell, A., Mishkin, P., Clark, J., et~al.: {Learning transferable visual models from natural language supervision}. In: ICML (2021)

\bibitem{Radford2018ImprovingLU}
Radford, A., Narasimhan, K.: {Improving Language Understanding by Generative Pre-Training} (2018)

\bibitem{reizenstein21co3d}
Reizenstein, J., Shapovalov, R., Henzler, P., Sbordone, L., Labatut, P., Novotny, D.: {Common Objects in 3D: Large-Scale Learning and Evaluation of Real-life 3D Category Reconstruction}. In: ICCV (2021)

\bibitem{ren2024grounded}
Ren, T., Liu, S., Zeng, A., Lin, J., Li, K., Cao, H., Chen, J., Huang, X., Chen, Y., Yan, F., Zeng, Z., Zhang, H., Li, F., Yang, J., Li, H., Jiang, Q., Zhang, L.: {Grounded SAM: Assembling Open-World Models for Diverse Visual Tasks} (2024)

\bibitem{rombach2021highresolution}
Rombach, R., Blattmann, A., Lorenz, D., Esser, P., Ommer, B.: {High-Resolution Image Synthesis with Latent Diffusion Models} (2021)

\bibitem{schuhmann2022laion}
Schuhmann, C., Beaumont, R., Vencu, R., Gordon, C., Wightman, R., Cherti, M., Coombes, T., Katta, A., Mullis, C., Wortsman, M., et~al.: {Laion-5b: An open large-scale dataset for training next generation image-text models}. NeurIPS  (2022)

\bibitem{8580422}
Shao, T., Yang, Y., Weng, Y., Hou, Q., Zhou, K.: {H-CNN: Spatial Hashing Based CNN for 3D Shape Analysis}. TVCG  (2020)

\bibitem{shen2021dmtet}
Shen, T., Gao, J., Yin, K., Liu, M.Y., Fidler, S.: {Deep Marching Tetrahedra: a Hybrid Representation for High-Resolution 3D Shape Synthesis}. In: NeurIPS (2021)

\bibitem{shi2020improving}
Shi, Z., Zhou, X., Qiu, X., Zhu, X.: {Improving image captioning with better use of captions}. CoRR  (2020)

\bibitem{song2016ssc}
Song, S., Yu, F., Zeng, A., Chang, A.X., Savva, M., Funkhouser, T.: {Semantic Scene Completion from a Single Depth Image}. CVPR  (2017)

\bibitem{su2021roformer}
Su, J., Lu, Y., Pan, S., Wen, B., Liu, Y.: {RoFormer: Enhanced Transformer with Rotary Position Embedding}. In: ICLR (2020)

\bibitem{varley2017shape}
Varley, J., DeChant, C., Richardson, A., Ruales, J., Allen, P.: {Shape completion enabled robotic grasping}. In: IROS (2017)

\bibitem{Wang2023OctFormer}
Wang, P.S.: {OctFormer: Octree-based Transformers for {3D} Point Clouds}. SIGGRAPH  (2023)

\bibitem{10.1145/3072959.3073608}
Wang, P.S., Liu, Y., Guo, Y.X., Sun, C.Y., Tong, X.: {O-CNN: Octree-Based Convolutional Neural Networks for 3D Shape Analysis}. SIGGRAPH  (2017)

\bibitem{PSWang2020}
Wang, P.S., Liu, Y., Tong, X.: {Deep Octree-based CNNs with Output-Guided Skip Connections for 3D Shape and Scene Completion}. In: CVPRW (2020)

\bibitem{Watson2022NovelVS}
Watson, D., Chan, W., Martin-Brualla, R., Ho, J., Tagliasacchi, A., Norouzi, M.: {Novel View Synthesis with Diffusion Models}. CoRR  (2022)

\bibitem{williams2021nkf}
Williams, F., Gojcic, Z., Khamis, S., Zorin, D., Bruna, J., Fidler, S., Litany, O.: {Neural Fields as Learnable Kernels for 3D Reconstruction}. In: CVPR (2022)

\bibitem{wu2023multiview}
Wu, C.Y., Johnson, J., Malik, J., Feichtenhofer, C., Gkioxari, G.: {Multiview Compressive Coding for 3{D} Reconstruction}. In: CVPR (2023)

\bibitem{wu2022point}
Wu, X., Lao, Y., Jiang, L., Liu, X., Zhao, H.: {Point transformer V2: Grouped Vector Attention and Partition-based Pooling}. In: NeurIPS (2022)

\bibitem{xiang2018posecnn}
Xiang, Y., Schmidt, T., Narayanan, V., Fox, D.: {PoseCNN: A Convolutional Neural Network for 6D Object Pose Estimation in Cluttered Scenes} (2018)

\bibitem{Xie2016}
Xie, S., Girshick, R., Dollár, P., Tu, Z., He, K.: {Aggregated Residual Transformations for Deep Neural Networks}. CVPR  (2017)

\bibitem{xu2022odise}
Xu, J., Liu, S., Vahdat, A., Byeon, W., Wang, X., De~Mello, S.: {ODISE: Open-Vocabulary Panoptic Segmentation with Text-to-Image Diffusion Models}. CVPR  (2023)

\bibitem{yan2022shapeformer}
Yan, X., Lin, L., Mitra, N.J., Lischinski, D., Cohen-Or, D., Huang, H.: {ShapeFormer: Transformer-based Shape Completion via Sparse Representation}. In: CVPR (2022)

\bibitem{yu2021pointr}
Yu, X., Rao, Y., Wang, Z., Liu, Z., Lu, J., Zhou, J.: {PoinTr: Diverse Point Cloud Completion with Geometry-Aware Transformers}. In: ICCV (2021)

\bibitem{zhai2021scaling}
Zhai, X., Kolesnikov, A., Houlsby, N., Beyer, L.: {Scaling vision transformers}. CVPR  (2022)

\bibitem{zhang2022cgca}
Zhang, D., Choi, C., Park, I., Kim, Y.M.: {Probabilistic Implicit Scene Completion}. In: ICLR (2022)

\bibitem{zhang2022glipv2}
Zhang, H., Zhang, P., Hu, X., Chen, Y.C., Li, L.H., Dai, X., Wang, L., Yuan, L., Hwang, J.N., Gao, J.: {GLIPv2: Unifying Localization and Vision-Language Understanding}. CoRR  (2022)

\bibitem{Zhang_2019_ICCV}
Zhang, P., Liu, W., Lei, Y., Lu, H., Yang, X.: {Cascaded Context Pyramid for Full-Resolution 3D Semantic Scene Completion}. In: ICCV (2019)

\bibitem{zhao2021point}
Zhao, H., Jiang, L., Jia, J., Torr, P.H., Koltun, V.: {Point transformer}. In: ICCV (2021)

\bibitem{zhu2017semantic}
Zhu, Y., Tian, Y., Mexatas, D., Doll{\'a}r, P.: {Semantic Amodal Segmentation}. In: CVPR (2017)

\end{thebibliography}

\title{Supplementary Material:\\ Zero-Shot Multi-Object Scene Completion}

\author{
Shun Iwase\inst{1,2} \and
Katherine Liu\inst{2} \and
Vitor Guizilini\inst{2} \and
Adrien Gaidon\inst{2} \and \\
Kris Kitani\inst{1,\star} \and
Rareș Ambruș\inst{2,\star} \and
Sergey Zakharov\inst{2,}\thanks{Equal advising.}
}

\authorrunning{S.~Iwase et al.}

\institute{Carnegie Mellon University \and Toyota Research Institute}

\maketitle

\section{Implementation Details of Baselines}
For occupancy-based networks such as AICNet~\cite{Li2020aicnet}, {ConvONet~\cite{Peng2020ECCV}}, POCO~\cite{Boulch_2022_CVPR}, and VoxFormer~\cite{li2023voxformer}, we use only the averaged BCE loss at LoD-$6$ ($L_{occ}^6$ in the main paper) for training. For surface-based methods such as MinkowskiNet~\cite{choy20194d} and OCNN~\cite{10.1145/3072959.3073608}, exact the same loss function as our method is used. We use the same hyperparameters for Adam~\cite{kingma:adam} as the proposed method for training.

\paragraph{VoxFormer~\cite{li2023voxformer}.}
We use the implementation from \url{https://github.com/NVlabs/VoxFormer}. We make a single modification to adapt to multi-object scene completion. Unlike the original setting, a measured depth map is more accurate than an estimated one. Thus, we directly use the input depth map to extract query tokens in Stage $1$. In addition, we leverage trilinear interpolation to reconstruct a surface at LoD-$7$.

\paragraph{ShapeFormer~\cite{yan2022shapeformer}.}
We borrow the code from \url{https://github.com/QhelDIV/ShapeFormer}. We train VQDIF and ShapeFormer following the paper, and pick the first prediction for evaluation.

\paragraph{MCC~\cite{wu2023multiview}.}
We choose the implementation from \url{https://github.com/facebookresearch/MCC}. We train the model with the lower number of sampling points being $1,100$ (twice more than the original implementation) due to their memory-expensive Transformer architecture. 

\paragraph{ConvONet~\cite{Peng2020ECCV}}
We use the implementation from \url{https://github.com/autonomousvision/convolutional_occupancy_networks}. We modify the network to accept the encoded feature from an RGB image as well as the point features through concatenation for a fair comparison.

\paragraph{POCO~\cite{Boulch_2022_CVPR}}
We choose the implementation from \url{https://github.com/valeoai/POCO}. As well as ConvONet~\cite{Peng2020ECCV}, we modify the network to accept the encoded feature from an RGB image as well as the point features through concatenation for a fair comparison.

\paragraph{AICNet~\cite{Li2020aicnet}.}
The implementation is borrowed from \url{https://github.com/waterljwant/SSC}. Since AICNet~\cite{Li2020aicnet} takes the same input as our method except a foreground mask. We only make a change in its output channel size from the number of classes to $2$ for occupancy prediction.

\paragraph{MinkowskiNet~\cite{choy20194d}.}
We adopt the implementation from \url{https://github.com/NVIDIA/MinkowskiEngine}. We use the network depth of $5$ (LoD-$9$ to LoD-$4$) for a fair comparison with the other networks. The occupancy probability of $0.5$ is also used for pruninig at each LoD.

\paragraph{OCNN~\cite{10.1145/3072959.3073608}}
The implementation is taken from \url{https://github.com/octree-nn/ocnn-pytorch}. We use the same network architecture and pruning strategy as MinkowskiNet~\cite{choy20194d} and our method. The sparse tensor structures are the key difference between OCNN~\cite{10.1145/3072959.3073608} and MinkowskiNet~\cite{choy20194d}. Specifically, MinkowskiNet and OCNN use the hash table and octree respectively.

\section{Evaluation Metrics}
To compute the metrics, we uniformly sample $100,000$ points on a surface for occupancy-based methods. For surface-based methods, we simply use the point locations predicted as occupied. Here, the predicted and ground-truth points are denoted as $\mathbf{P}_{\text{pd}}$ and $\mathbf{P}_{\text{gt}}$ respectively.

\paragraph{Chamfer distance (CD).}

The Chamfer distance $\text{CD}(\mathbf{P}_{\text{pd}}, \mathbf{P}_{\text{gt}})$ is expressed as

\begin{equation}
\begin{aligned}
\text{CD}(\mathbf{P}_{\text{pd}}, \mathbf{P}_{\text{gt}}) &= \frac{1}{2|\mathbf{P}_{\text{pd}}|} \sum_{\mathbf{x}_{\text{pd}} \in \mathbf{\mathbf{P}_{\text{pd}}} } \min_{\mathbf{x}_{\text{gt}} \in \mathbf{\mathbf{P}_{\text{gt}}}} ||\mathbf{x}_{\text{pd}} - \mathbf{x}_{\text{gt}}|| \\
&+ \frac{1}{2|\mathbf{P}_{\text{gt}}|} \sum_{\mathbf{x}_{\text{gt}} \in \mathbf{\mathbf{P}_{\text{gt}}} } \min_{\mathbf{x}_{\text{pd}} \in \mathbf{\mathbf{P}_{\text{pd}}}} ||\mathbf{x}_{\text{gt}} - \mathbf{x}_{\text{pd}}||.
\end{aligned}
\end{equation}

\paragraph{F-1 score.}
The F-1 score is computed by
\begin{equation}
\begin{aligned}
P & =\frac{\left|\left\{\mathbf{x}_{\mathrm{pd}} \in \mathbf{P}_{\mathrm{pd}} \mid \min_{\mathbf{x}_{\mathrm{gt}} \in \mathbf{P}_{ \mathrm{gt}}}\left\|\mathbf{x}_{\mathrm{gt}}-\mathbf{x}_{\mathrm{pd}}\right\|<\eta\right\}\right|}{\left|\mathbf{P}_{\mathrm{pd}}\right|}, \\
R & =\frac{\left|\left\{\mathbf{x}_{\mathrm{gt}} \in \mathbf{P}_{\mathrm{gt}}\mid \min_{\mathbf{x}_{\mathrm{pd}} \in \mathbf{P}_{\mathrm{pd}}} \left\| \mathbf{x}_{\mathrm{pd}}-\mathbf{x}_{\mathrm{gt}} \right\|<\eta\right\}\right|}{\left|\mathbf{P}_{\mathrm{gt}}\right|}, \\
\end{aligned}
\end{equation}

\begin{equation}
    \text{F-1} = \frac{2 P R}{P + R}.
\end{equation}
where we set $\eta$ to $10$ mm for all the experiments.

\paragraph{Normal consistency (NC).}
Normal consistency measures the alignment of normals between the predicted and ground surfaces.
\begin{equation}
\text{NC}(\mathbf{N}_{\text{pd}}, \mathbf{N}_{\text{gt}}) = \frac{1}{2|\mathbf{N}_{\text{pd}}|} \sum_{\mathbf{n}_{\text{pd}} \in \mathbf{\mathbf{N}_{\text{pd}}}} \left(\mathbf{n}_{\text{pd}} \cdot \mathbf{n}^{*}_{\text{gt}}\right) + \frac{1}{2|\mathbf{N}_{\text{gt}}|} \sum_{\mathbf{n}_{\text{gt}} \in \mathbf{\mathbf{N}_{\text{gt}}} } \left(\mathbf{n}_{\text{gt}} \cdot \mathbf{n}^{*}_{\text{pd}}\right).
\end{equation}
where $\mathbf{n}^{*}_{\text{gt}}$ and $\mathbf{n}^{*}_{\text{pd}}$ refer the nearest normal vectors.

\section{Derivation of RoPE~\cite{su2021roformer}}
RoPE~\cite{su2021roformer} utilizes a rotation matrix to encode positional information to features. Given normalized 1D axial coordinate $x \in \mathbb{R}$, $R: \mathbb{R} \rightarrow \mathbb{R}^{\left\lfloor{D^{\prime} / 3}\right\rfloor \times \left\lfloor{D^{\prime} / 3}\right\rfloor}$ is defined as
\begin{equation}
    R (x) = \begin{bmatrix}
        \cos x \theta_1 & - \sin x \theta_1 & \cdots & 0 & 0 \\
        \sin x \theta_1 & \cos x \theta_1 & \cdots & 0 & 0 \\
        \vdots & \vdots & \ddots & \vdots & \vdots  \\
        0 & 0 & \cdots & \cos x \theta_{k/2} & - \sin x \theta_{k/2} \\
        0 & 0 & \cdots & \sin x \theta_{k/2} & \cos x \theta_{k/2} \\
    \end{bmatrix},
\end{equation}
where $\theta_i = \left(1 + \frac{ \left\lfloor{D^\prime / 2 }\right\rfloor - 1}{\left\lfloor{D^\prime / 6 }\right\rfloor - 1} \right) \left(i - 1 \right)\pi, i \in \left[1,2, \cdots, \left\lfloor{D^\prime / 6 }\right\rfloor \right]$.

\begin{equation}
    \mathbf{R}_i = \begin{bmatrix}
         R (p^x_i) & \mathbf{0} & \mathbf{0} & \mathbf{0} \\
         \mathbf{0} & R(p^y_i) & \mathbf{0} & \mathbf{0} \\
         \mathbf{0} & \mathbf{0} & R(p^y_i) & \mathbf{0} \\
         \mathbf{0} & \mathbf{0} & \mathbf{0} & \mathbf{I}
    \end{bmatrix} \in \mathbb{R}^{D^\prime \times D^\prime},
\end{equation}

\begin{equation}
    \mathbf{f}^{\prime}_i =  \mathbf{R}_i \mathbf{f}_i,
\end{equation}
where $\mathbf{f}_i \in \mathbb{R}^{D^\prime}$, and $\mathbf{p}_i \in \mathbb{R}^{3}$ is an $i$-th octree feature and coordinates. 

\section{Additional Experiments}

\begin{table}[h]
    \centering
    \resizebox{\columnwidth}{!}{
\begin{tabular}{c||c|c||c|c|c|c|c}
& \#Obj & Mask Source & CD$\downarrow$ &$\text{CD}_{\text{occ}}$ $\downarrow$ & F1$\uparrow$ & NC$\uparrow$ & Runtime$\downarrow$ \\ \hline
MCC~\cite{wu2023multiview}       & $1$ & Ground truth & $14.39$ & $20.26$ & $0.482$ & $0.694$ & $3.5 \times 10^4$ \\
ZeroShape~\cite{huang2023zeroshape} & $1$ & Ground truth & $12.19$ & $17.44$ & $0.603$ & $0.703$ & $965.5$ \\ \hline
\multirow{5}{*}{Ours} & $1$ & G-SAM~\cite{ren2024grounded}      & $14.98$ & $20.73$ & $0.666$ & $0.683$ & $240.3$ \\
                      & $N$ & G-SAM~\cite{ren2024grounded}      & $13.56$ & $16.19$ & $0.700$ & $0.699$ & $83.4$ \\
                      & $N$ & InSPyReNet~\cite{kim2022revisiting} & $12.31$ & $14.47$ & $0.724$ & $0.712$ & $84.0$ \\
                      & $1$ & Ground truth         & $8.55$ & $12.48$ & $0.758$ & $0.730$ & $248.3$ \\
                      & $N$ & Ground truth         & $\textbf{6.97}$ & $\textbf{9.31}$  & $\textbf{0.803}$ & $\textbf{0.750}$ & $\textbf{85.1}$ \\
    \end{tabular}}
    \caption{Ablations of mask sources and the number of target objects on the HOPE dataset.}
    \label{tab:ablation1}
\end{table}

\subsection{Comparison against single-object methods}

We trained our method, MCC~\cite{lin2021fusion} and ZeroShape~\cite{huang2023zeroshape}
on our synthetic dataset with a single-object setup. For a fair comparison, we use ground-truth camera intrinsics and depth maps for Zeroshape. During evaluation, we complete each object individually and then concatenate all the completed objects in a scene.
\Cref{tab:ablation1} demonstrates that our method with single- and multi-object setups outperforms the others regarding completion quality and runtime.
Here, $1$ and $N$ in \#Obj denote single- and multi-object setups, and $\text{CD}_{occ}$ is Chamfer distance of occluded surfaces. 
The large difference between $\text{CD}$ and $\text{CD}_{occ}$ of single-object methods clearly show its poor occlusion handling due to the lack of multi-object reasoning.

\subsection{Foreground vs Instance Masks}
Zero-shot 2D instance segmentation of cluttered scenes is still challenging. For instance, the SoTA foreground detection model (InSPyReNet~\cite{kim2022revisiting}) gives a $14.9\%$ higher IoU of a foreground mask than Grounded-SAM~\cite{ren2024grounded} (G-SAM), the latest zero-shot instance segmentation model, on HOPE dataset ($69.3\%$ vs $54.4\%$). For G-SAM, foreground masks are computed by combining its instance mask predictions, and its input prompt are manually tuned to improve an IoU. Further, \Cref{tab:ablation1} validates that using foreground masks during inference largely improves the final completion quality.

\end{document}